\documentclass[letterpaper, 10 pt, journal, twoside]{IEEEtran}

\IEEEoverridecommandlockouts                              
\usepackage[pagebackref=true,breaklinks=true,letterpaper=true,colorlinks,bookmarks=false]{hyperref}
\usepackage{graphics} 
\usepackage{epsfig} 
\usepackage{mathptmx} 
\usepackage{times} 
\usepackage{amsmath} 
\usepackage{amssymb}  
\usepackage{caption}

\usepackage{amsmath,amsfonts,bm}

\def\eqref#1{equation~\ref{#1}}

\def\1{\bm{1}}

\DeclareMathAlphabet{\mathsfit}{\encodingdefault}{\sfdefault}{m}{sl}
\SetMathAlphabet{\mathsfit}{bold}{\encodingdefault}{\sfdefault}{bx}{n}

\usepackage{graphicx}
\usepackage{listings}
\usepackage{algorithm}
\usepackage{algpseudocode}
\usepackage[autostyle]{csquotes}
\usepackage{hyperref}
\usepackage{url}
\usepackage{pifont}
\usepackage{xcolor}
\usepackage{booktabs, multirow}
\usepackage{makecell}

\definecolor{darkgreen}{RGB}{0,100,0}
\definecolor{darkblue}{RGB}{0,0,100}
\definecolor{darkyellow}{RGB}{251, 236, 93}

\newcommand{\arxiv}[1]{}
\newcommand{\ral}[1]{#1}
\newcommand{\citep}[1]{\cite{#1}}
\newcommand{\citet}[1]{\cite{#1}}

\newcommand{\rbt}[1]{#1}
\newcommand{\finalrev}[1]{#1}

\begin{document}

\title{
Autoregressive Action Sequence Learning for Robotic Manipulation
}

\author{Xinyu Zhang, Yuhan Liu, Haonan Chang, Liam Schramm and  Abdeslam Boularias
\thanks{Manuscript received: November 18, 2024; Revised:
January 23, 2025; Accepted: February 25, 2025.}
\thanks{This paper was recommended for publication by
Editor Markus Vincze upon evaluation of the Associate Editor and Reviewers' comments. This work is partly
supported by NSF awards 1846043 and 2132972.}
\thanks{The authors are with the Department of Computer Science, Rutgers University, 
        {\tt\small xz653@rutgers.edu}. 
}%
\thanks{Digital Object Identifier (DOI): see top of this page.}
}

\markboth{IEEE Robotics and Automation Letters. Preprint Version. Accepted March, 2025}
{Zhang \MakeLowercase{\textit{et al.}}: Autoregressive Action Sequence Learning for Robotic Manipulation}

\maketitle

\begin{abstract}

Designing a universal policy architecture that performs well across diverse robots and task configurations remains a key challenge. In this work, we address this by representing robot actions as sequential data and generating actions through autoregressive sequence modeling. Existing autoregressive architectures generate end-effector waypoints sequentially as word tokens in language modeling, which are limited to low-frequency control tasks.
Unlike language, robot actions are heterogeneous and often include high-frequency continuous values---such as joint positions, 2D pixel coordinates, and end-effector poses—which are not easily suited for language-based modeling. Based on this insight, we extend causal transformers' single-token prediction to support predicting a variable number of tokens in a single step through our Chunking Causal Transformer (CCT). 
This enhancement enables robust performance across diverse tasks of various control frequencies, greater efficiency by having fewer autoregression steps, and 
lead to a hybrid action sequence design by mixing different types of actions and using a different chunk size for each action type. 
Based on CCT, we propose the Autoregressive Policy (ARP) architecture, which solves manipulation tasks by generating hybrid action sequences. 
We evaluate ARP across diverse robotic manipulation environments, including Push-T, ALOHA, and RLBench, and show that ARP, as a universal architecture, \rbt{matches or} outperforms the environment-specific state-of-the-art in all tested benchmarks, while being more efficient in computation and parameter sizes. Videos of our real robot demonstrations, all source code and the pretrained models of ARP can be found at \texttt{\url{http://github.com/mlzxy/arp}}.

\end{abstract}


\section{Introduction}
\label{sec:introduction}

Autoregressive models are the basis of recent breakthroughs in natural language processing~\citep{min2023recent}. These models predict the next token in a sequence based on the previous tokens. Autoregressive models are typically implemented as causal transformers, where each token attends only to preceding ones, and they are trained with the single objective of maximizing the conditional likelihood of each token. Despite their simplicity, autoregressive models such as GPTs~\citep{mann2020gpt3} are shown to demonstrate a reasoning ability that can capture causal dependencies~\citep{prystawski2024stepbystep}. In this work, we present a new universal autoregressive architecture that can be used for various robot manipulation tasks in diverse environments.

Decision Transformer (DT) and Trajectory Transformer (TT) are two pioneering approaches that use autoregressive models to solve control tasks~\citep{chen2021decisionT, janner2021trajT}. These methods learn to generate trajectories as $(R_1, s_1, a_1, \dots, R_T, s_T, a_T)$, where $R_t, s_t, a_t$ respectively denote the reward-to-go~\citep{tamar2016rewardtogo}, the state, and the action at time-step $t$. However, these methods are primarily applied to tasks with fully observed, low-dimensional states---which is rarely the case in robotics applications. Recent work focuses on applying autoregression only on action sequences, such as Gato~\citep{reed2022gato}, VIMA~\citep{jiang2022vima}, ManipLLM~\citep{li2024manipllm}. Despite their impressive results, these methods remain limited to low-frequency control tasks, as they represent robot actions with key end-effector waypoints and generate one action at a time, similar to word generation in language modeling.


However, unlike language, robot actions are heterogeneous and include continuous values---such as joint positions, 2D pixel coordinates, and effector poses. Additionally, in high-frequency control tasks, continuous actions are expected to maintain temporal smoothness---a requirement absent in language modeling.
To adapt autoregressive models for robotic tasks, we propose the Chunking Causal Transformer (CCT), an improved version of the causal transformer used in standard autoregressive models. CCT introduces an important modification: it predicts the future tokens (a chunk of actions) from empty tokens rather than from the original sequence, as illustrated in Figure~\ref{fig:teacher-forcing}. In doing so, CCT extends the next-single-token prediction of causal transformer to \textit{chunking autoregression}---next-multi-token prediction in a single step. Despite its simplicity, chunking autoregression offers three key advantages:

\begin{enumerate}
    \item  Predicting multiple temporally correlated actions in a single step addresses the primary limitation of autoregressive models in high-frequency control tasks.
    \item Chunking autoregression increases efficiency by reducing the number of inference passes that are required.
    \item Variable chunk sizes enable a flexible action sequence design, such as mixing different types of actions and using a different chunk size for each type. For example, high-level actions, like sparse 2D waypoints, can be predicted first sequentially to guide the prediction of low-level actions, such as joint positions, in larger chunks. We show action sequence designs of our real robot task, Push-T, ALOHA, and RLBench in Figure~\ref{fig:seq-design}. 
\end{enumerate}

\begin{figure*}[t]
    \centering 
    \includegraphics[width=\linewidth]{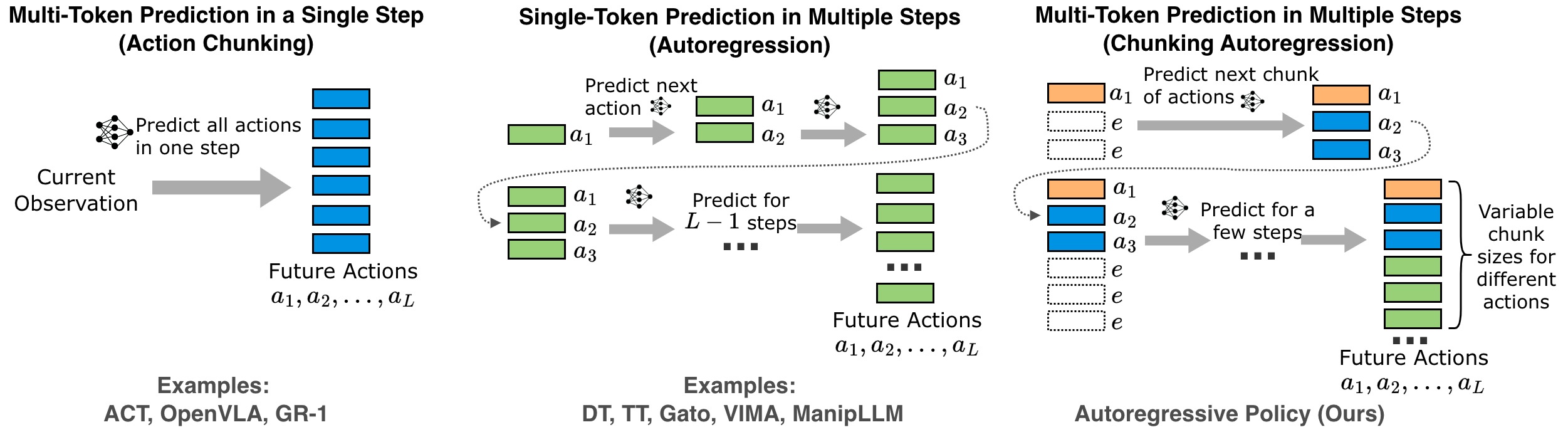}
    \arxiv{\vspace{-2em}}
    \ral{\vspace{-1.5em}}
    \caption{\textbf{Existing Works versus Our Autoregressive Policy.} 
    Action chunking models (left) predict all action tokens in a single step~\citep{zhao2023aloha, kim2024openvla, wu2023gr1}. Standard autoregression models (middle) generate one action token in each step, which is inefficient and unsuitable for high-frequency control tasks~\cite{chen2021decisionT, janner2021trajT, reed2022gato, jiang2022vima, li2024manipllm}. Our proposed chunking autoregression (right) generates a chunk of variable number of action tokens per step,
    offering greater efficiency, strong performance across diverse tasks, and flexibility in designing hybrid action sequences. We compare the performance of these three action prediction strategies in Figure~\ref{tab:generation-mode}.
    Note all models use Model Predictive Control to predict $L$ actions, execute them, update the observation, and then predict actions again. 
    Autoregressive generation is performed without executing actions or changing the current observation. 
    }
    \label{fig:comparison}
    \vspace{-1em}
\end{figure*}

While action chunking has already been introduced in the Action Chunking Transformer (ACT) by \citet{zhao2023aloha}, ACT is a one-step prediction model with a fixed chunk size. Instead, our approach supports variable chunk sizes for generating hybrid action sequences. 
Figure~\ref{tab:cross-env} shows that our method outperforms ACT by a significant margin in all environments. Figure~\ref{tab:generation-mode} shows that our chunking autoregression is the key factor behind this strong performance. We illustrate the essential difference between action chunking, standard autoregression, and our chunking autoregression in Figure~\ref{fig:comparison}. We illustrates why chunking autoregression works in Fig.~\ref{fig:chunking-auto-regression}.

\begin{figure}[ht]
    \centering
     \vspace{-1.3em}
    \includegraphics[width=1.02\linewidth]{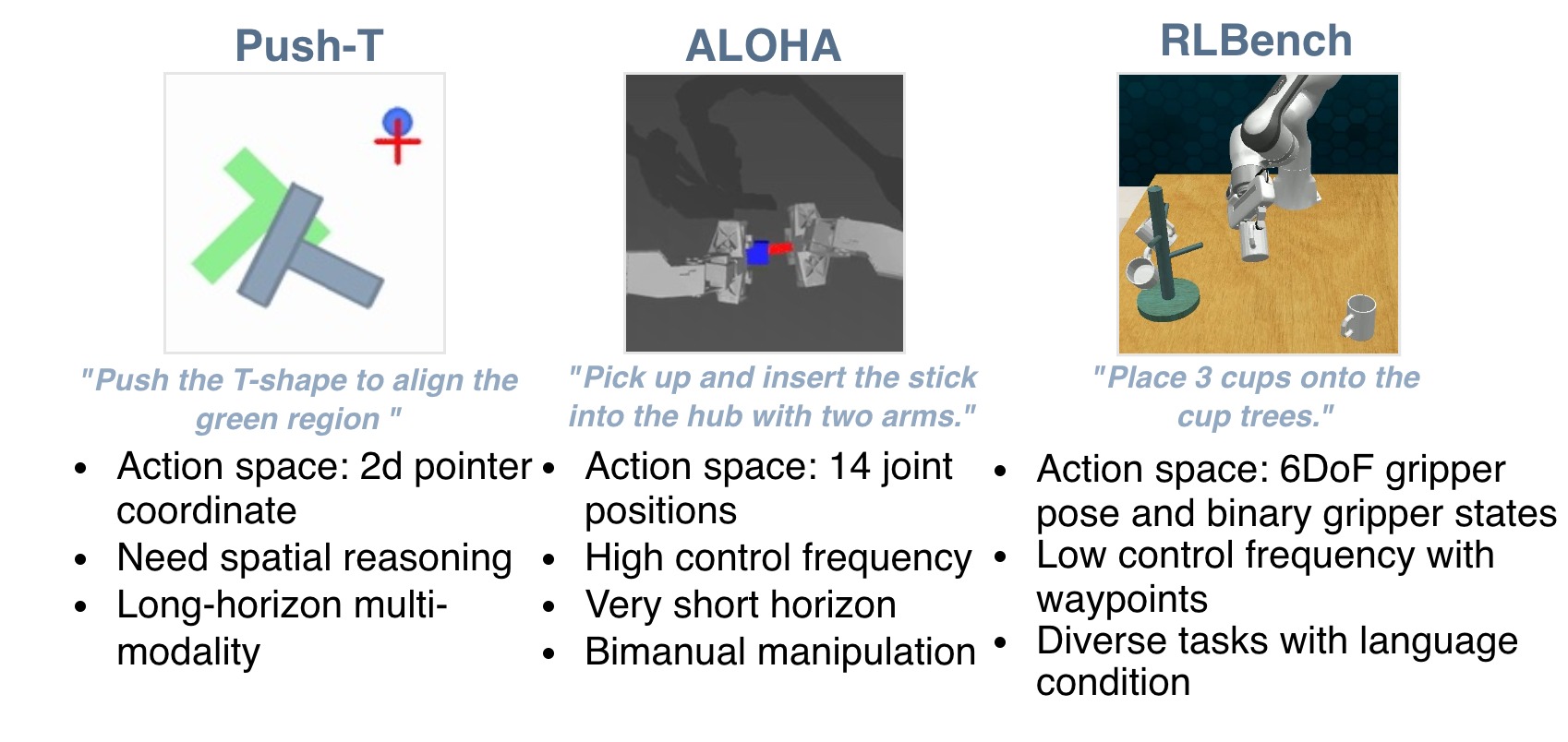}
    \vspace{-2.5em}
    \caption{\textbf{Overview of the simulation environments.} We evaluate our method on Push-T, ALOHA, and RLBench, three task suites with significantly different properties and requirements. Push-T~\citep{chi2023diffp} requires many steps to complete (long horizon) and where same sub-goals can be reached in various ways (multi-modality). 
    ALOHA~\citep{zhao2023aloha} has a high-dimensional action space (14 joints of two robot arms), a high control frequency (50Hz), and a short time limit (8 secs). RLBench~\citep{james2020rlbench} has only the  gripper pose as action but contains 18 different language-conditioned tasks.
    }
    \label{fig:env-intro}
\end{figure}

To summarize, our contributions are threefold. (1) We propose the Chunking Causal Transformer (CCT), which extends the single-token prediction of causal transformer to multi-token prediction, and therefore enables chunking autoregression. We also design a novel attention interleaving strategy that allows CCT to be trained efficiently with teacher-forcing, as shown in Figure~\ref{fig:teacher-forcing}. 
(2) Based on our CCT, we present the Auto-regressive Policy (ARP), a model that learns to generate heterogeneous action sequences autoregressively for solving robotic manipulation tasks. The ARP architecture is summarized in Figure~\ref{fig:main}. (3) We evaluate the same ARP architecture across Push-T~\citep{chi2023diffp}, ALOHA~\citep{zhao2023aloha}, and RLBench~\citep{james2020rlbench}, three environments with diverse manipulation tasks and control modalities, as outlined in Figure~\ref{fig:env-intro}. Our study shows that ARP \rbt{matches or} outperforms all environment-specific SoTAs, while being more efficient computationally and using smaller parameter sizes, as in Figure~\ref{fig:main-result}. In addition, we evaluate ARP with a real robot on a challenging, contact-rich nut-tightening task, as in Figure~\ref{fig:real-robot}.

\section{Related Work}
\label{sec:relatedwork}

\textbf{Learning robotic manipulation from demonstrations.} Imitation learning enables robots to learn to perform tasks demonstrated by experts~\citep{zhang2024oneshot, zare2024ilsurvey}. Recently, 
various methods have been developed for manipulation learning with different task constraints and control modalities. 
Notably, \citet{chi2023diffp} proposed the Diffusion Policy (DP)  method for solving the Push-T task.
\citet{zhao2023aloha} proposed the Action Chunking Transformer (ACT) for bi-manual manipulation tasks in the ALOHA environment. 
\citet{goyal2024rvt2} proposed RVT-2 for language-conditioned tasks in the RLBench environment~\citep{james2020rlbench}. 
We outline these environments and the corresponding state-of-the-art (SoTA) solutions in Figure~\ref{fig:env-intro} and Figure~\ref{fig:existing-sota}, respectively. 
In contrast, our proposed autoregressive policy is a universal architecture that outperforms each environment-specific SoTA on Push-T, ALOHA, and RLBench.

\textbf{Autoregressive models for control tasks.} Besides the pioneering Decision Transformer and Trajectory Transformer, recent works such as VIMA~\citep{jiang2022vima}, Gato~\citep{reed2022gato}, GR1~\citep{wu2023gr1}, OpenVLA~\citep{kim2024openvla} and ManipLLM~\citep{li2024manipllm} have looked into designing autoregressive models for robotic tasks. 
Despite the impressive results, these approaches are limited to low-frequency control tasks that rely on end-effector waypoints~\citep{kim2024openvla}. 
GR1 and OpenVLA only use autoregression in their LLM backbones and do not apply autoregression to solve control tasks. 
Most of the existing works require fine-tuning a large language model (LLM) such as LLaMA~\citep{li2024manipllm} to include target end-effector poses within text-based responses or predict poses from LLM's hidden features. 
The reliance on resource-intensive LLMs leads to large computational overhead, even for tasks that could be addressed with lightweight models.
Without these constraints, our autoregressive policy outperforms SoTAs in multiple environments while being more efficient in computation and parameter sizes.

\begin{figure*}[t!]
    \centering
       \vspace{-1em}
    \includegraphics[width=0.8\linewidth]{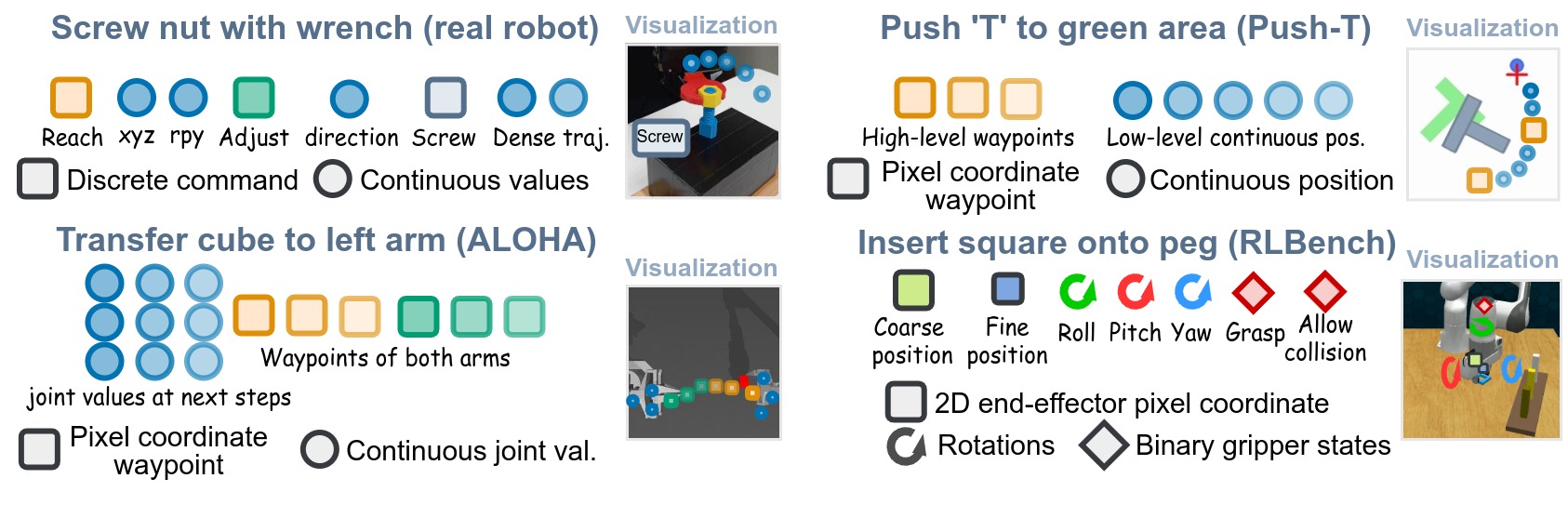}
    \vspace{-1em}
    \caption{\textbf{Learned Action Sequences.} 
    In Push-T, our model predicts a sequence of high-level waypoints, followed by a sequence of low-level positions that connect the waypoints together and form the pushing trajectory, analogous to hierarchical planning~\citep{hafner2022hiera-plan}. 
    In ALOHA, we predict the joint values and then end-effector waypoints conditioned on the joint values, a process akin to forward kinematics~\citep{kucuk2006robot}. We bypass the waypoint generation during inference. 
    In RLBench, we predict the target end-effector's position first, then gripper rotation and state in that position. 
    For our real robot experiment, we define a set of primitive actions, as detailed in section~\ref{sec:real-robot}. We predict the action type and then continuous values of that action.
    }
    \label{fig:seq-design}
    \vspace{-1.0em}
\end{figure*}

\textbf{Hierarchical policies.} Planning actions on multiple levels of abstraction is an important ability~\citep{pateria2021hiera-survey}. Existing methods 
generally separate the designs of low-level and high-level policies, 
 and uses different modules for the different levels of abstraction~\citep{pateria2021hiera-survey, belkhale2023hydra}. This complicated procedure prohibits a wider application. In contrast, our autoregressive policy is able to capture the underlying causal dependencies in robotic tasks by predicting a sequence of actions of different levels of abstraction by using a single model, as shown by our action sequence designs for diverse environments in Figure~\ref{fig:seq-design}.

 %

\section{Method}
\label{sec:method}

\begin{figure}
    \centering
    \includegraphics[width=\linewidth]{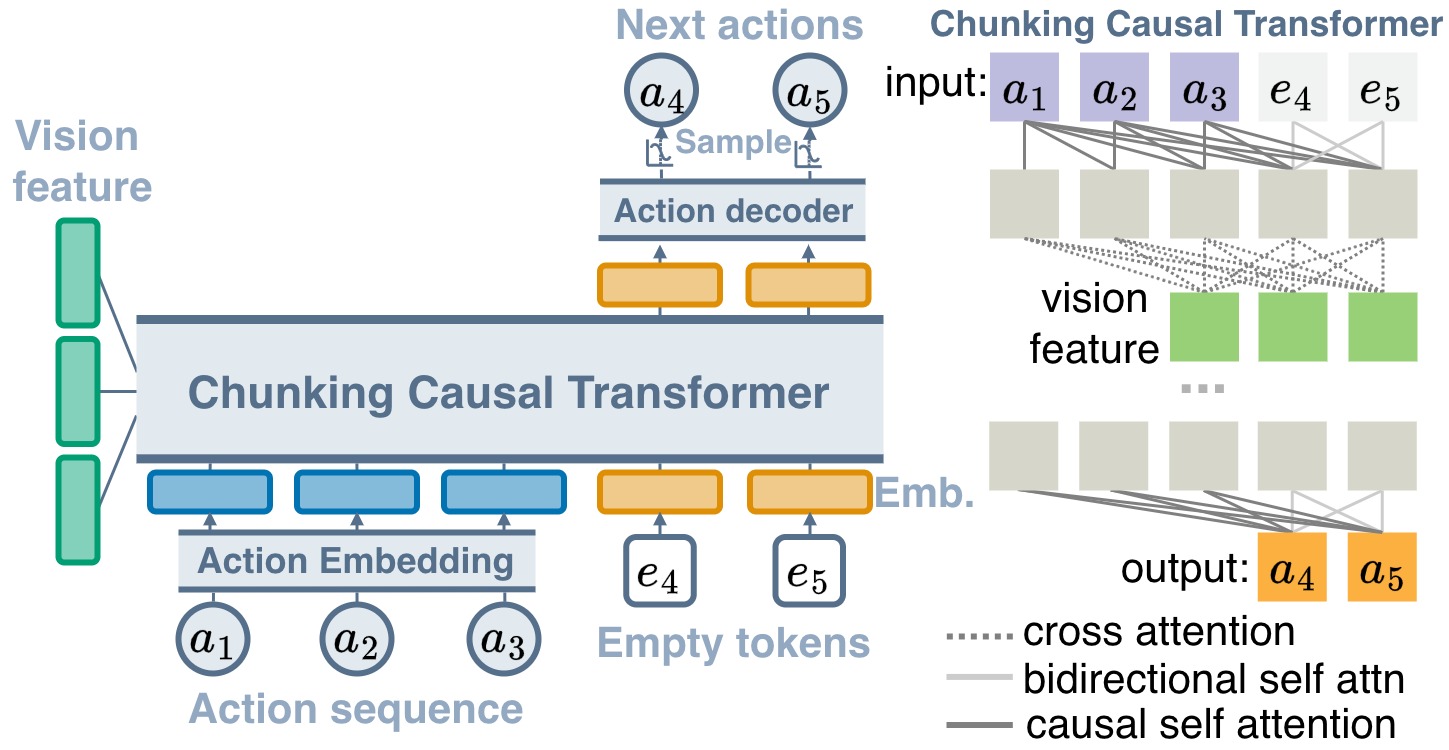}
    \vspace{-1.5em}
    \caption{\textbf{Autoregressive Policy Architecture.} 
    A sequence of past actions and a chunk of empty tokens are concatenated and projected into embeddings. Empty tokens correspond to future actions, which are unknown and need to be predicted. 
    These embeddings are fed into our Chunking Causal Transformer (CCT) along with the vision features of the current observation. 
    CCT alternates between self-attention within the sequence embeddings and cross-attention with the vision features. Self-attention is causal for the input actions and bidirectional among the empty tokens. 
    Distributions of future actions are decoded from the updated embeddings of the empty tokens.
    }
    \label{fig:main}
    \vspace{-1em}
\end{figure}

In this section, we present the Auto-regressive Policy (ARP), built upon our Chunking Causal Transformer (CCT), which generates robot action sequences autoregressively.
We summarize the architecture in Figure~\ref{fig:main}.

\textbf{Action sequence modeling.} Unlike natural language, robot actions lack a universal vocabulary. 
As shown in Figure~\ref{fig:env-intro}, different robot tasks may require drastically different types of actions.
Therefore, we represent actions as structured sequences whose formats are tailored for each family of tasks. Figure~\ref{fig:seq-design} showcases the formats of the action sequences generated in our real robot experiment, Push-T, ALOHA, and RLBench tasks.

\begin{figure}[h]
    \centering
    \includegraphics[width=\linewidth]{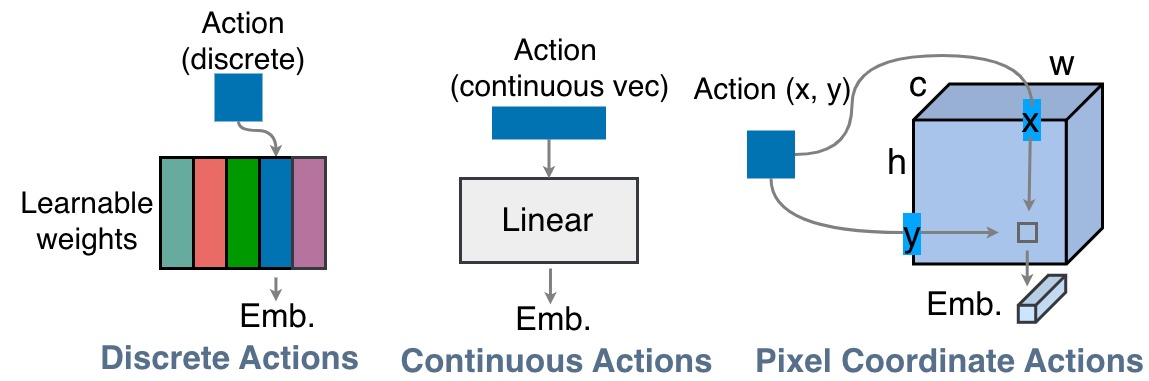}
    \vspace{-1.5em}
    \caption{\textbf{Embeddings for Discrete, Continuous, and Pixel-coordinate Actions.} 
    Discrete actions are embedded by a simple table lookup on a weight matrix. 
    Continuous actions are embedded with a linear layer. 
    Pixel-coordinate actions are embedded by retrieving the point-wise features at the coordinates on the visual feature maps. }
    \label{fig:embed-a}
    \vspace{-1em}
\end{figure}

\textbf{Embedding and decoding heterogeneous actions.} Language models map each word to a continuous vector called word embedding. 
The word embeddings of the input sentences are fed into a causal transformer. The distribution of the next word is decoded from the output embedding of the last word with a linear layer. Figure~\ref{fig:embed-a} and \ref{fig:decode-a} illustrate our embedding and decoding methods for robot actions.
Discrete actions are embedded by a table lookup on a weight matrix and decoded into a categorical distribution with a linear layer, similar to words in language modeling. Continuous actions are embedded with a linear layer and decoded into the parameters of a Gaussian mixture distribution with another linear layer. Actions that are defined as pixel coordinates are embedded by retrieving the point-wise features at the coordinates on a visual feature map. The output spatial distribution is obtained by multiplying the output embedding with the visual feature map, and converting the result into a 2-d heatmap with the up-sampling operator from RAFT~\citep{teed2020raft}.

\begin{figure*}[t!]
    \centering
    \vspace{-1em}
    \includegraphics[width=0.8\linewidth]{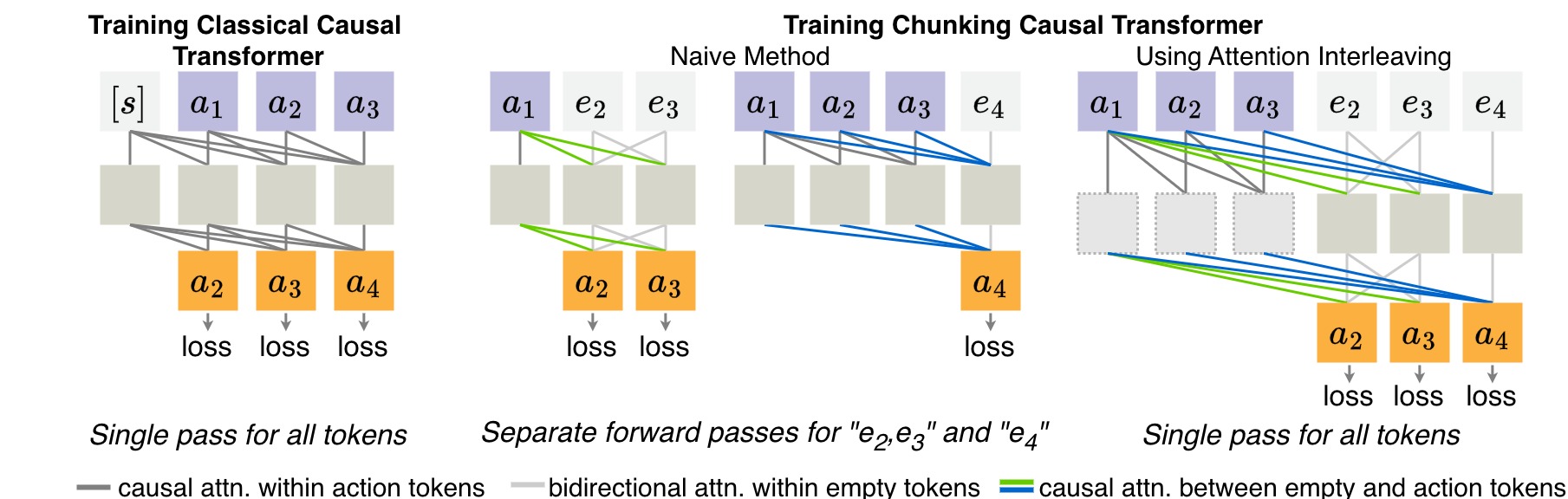}
    \caption{\textbf{Training Chunking Causal Transformer (CCT) with Teacher-forcing.} Causal transformers are trained efficiently with only a single forward pass for all tokens in a given sequence. However, suppose $a_2,a_3$ and $a_4$ are in separate chunks, the CCT forward passes of predicting $a_2,a_3$ and  $a_4$ cannot be merged directly. Naively running separate passes significantly increases computation costs, as in Figure~\ref{fig:attention-interleave}. 
    With the proposed attention interleaving, we can update all the empty tokens in a single pass, regardless of the number of tokens \rbt{or variable chunk size configurations}. \rbt{The key idea is to update empty tokens and action tokens separately and reuse the causally attended action tokens to update empty tokens.}
    A \rbt{step-by-step} example is provided in 
 \rbt{Figure~\ref{fig:attn-interleave-video} and } {\texttt{Video/attention-interleaving-tour.mp4}} in the supplementary.}
    \label{fig:teacher-forcing}
    \vspace{-1.5em}
\end{figure*}

\textbf{Chunking causal transformer.} Figure~\ref{fig:compare-cct-ct} illustrates the essential difference between a causal transformer and our CCT. A causal transformer modifies the token embedding with causal attention so that the last token becomes the next token. 
Our CCT modifies the token embedding with
causal attention for the action tokens $a_i$ and bidirectional attention for the empty tokens $e_i$ (future actions). The empty
tokens become the next tokens.
This allows the prediction of a chunk of variable number of next tokens at once in a single forward pass by adding empty tokens, which enables action chunking during autoregressive generation.
We study the impacts of our chunking autoregression in detail in Section~\ref{sec:experiment}.
In ARP, CCT alternates between self-attention within the input embeddings and cross-attention with vision features, as in Figure~\ref{fig:main}. We extract vision features from a standard backbone identical to the ones used in SoTA methods, as detailed in section~\ref{sec:experiment}.

\textbf{Train-time attention interleaving.} During training, a causal transformer is taught to predict each token in a given sequence by consuming all previous ground-truth tokens as input. This training strategy is named teacher-forcing~\citep{teacher-forcing}. As shown in Figure~\ref{fig:teacher-forcing}, only a single forward pass is required for training samples such as $a_1, a_2, a_3 \rightarrow a_4$ (predict $a_4$ from $a_1, a_2, a_3$),  $a_1, a_2 \rightarrow a_3$, and  $a_1 \rightarrow a_2$.
Causal transformers are therefore efficiently trained with teacher-forcing. 
We follow this teacher-forcing strategy. 
However, training CCT yields separate forward passes per chunk. For example, the prediction of $a_4$ depends on $a_2, a_3$, as in $a_1, a_2, a_3, e_4 \rightarrow a_4$, but $a_2, a_3$ need to be replaced with $e_2, e_3$ to predict them from $a_1$, as in $a_1, e_2, e_3 \rightarrow a_2, a_3$. This prohibits the use of a single forward pass for both $a_1, a_2, a_3, e_4 \rightarrow a_4$ and $a_1, e_2, e_3 \rightarrow a_2, a_3$. Note $a_i$ denotes the $i$-th action and $e_i$ denotes the empty token for $i$-th action.
This issue increases the training cost and complicates the training procedure.  


To resolve this, \rbt{we have all action tokens and their corresponding empty tokens as model input, as in \ding{202} \ding{203} of Figure~\ref{fig:attn-interleave-video}. We then compute bidirectional attention within empty tokens and causal attention from empty tokens to action tokens using two masked attentions, as in \ding{204} \ding{205}. 
Next, we compute causal attention within action tokens, as in \ding{206}. This leverages the fact that action tokens are independent of future empty tokens. As a result, the updated action tokens are computed once and reused in the next layer, as in \ding{207}.}
This enables a single forward pass of all tokens in three attention operations, regardless of the number of tokens \rbt{or variable chunk configurations.}
We name this procedure \textit{attention interleaving}.
Figure~\ref{fig:attention-interleave} demonstrates the reduced MACs of training with attention interleaving.
We implement attention interleaving as an internal acceleration mechanism of the transformer layer, which is transparent to other network modules.
Note that attention interleaving is only used during training and incurs no additional inference cost.

\begin{figure}[h]
    \centering
    \includegraphics[width=\linewidth]{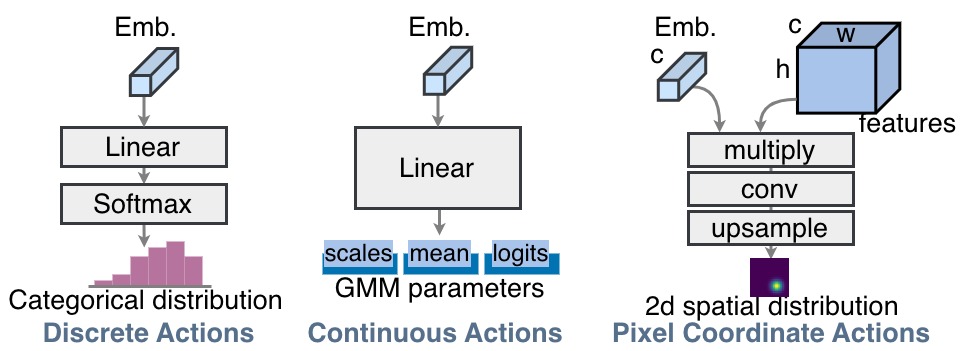}
    \vspace{-1.5em}
    \caption{\textbf{Decoders for Discrete, Continuous, and Pixel-coordinate Actions.} For discrete actions, we decode the action embeddings into a categorical distribution with a linear layer followed by a softmax operation. For continuous actions, we decode the embeddings into the parameters of a Gaussian mixture distribution with a linear layer. For the pixel-coordinate actions, we multiply the embedding with a visual feature map or a weight tensor, and convert the result into a 2-d heatmap.}
    \label{fig:decode-a}
    \vspace{-1em}
\end{figure}

\begin{figure}
    \centering
     \resizebox{\linewidth}{!}{
    \begin{tabular}{c|c|rr|cc}\toprule
Environment &\multicolumn{1}{c|}{Method}  &\multicolumn{2}{c|}{Success Rate} &\#MACs &\#Params \\\midrule
\multirow{2}{*}{Push-T} & \multicolumn{1}{c|}{Diffusion Policy}  &\multicolumn{2}{c|}{78.8} &6.8G &25.5M \\
&\multicolumn{1}{c|}{ARP (Ours)} &\multicolumn{2}{c|}{\textbf{87.1}} &\textbf{3.7G} &\textbf{23.5M} \\\midrule
\multirow{2}{*}{ALOHA} &\multicolumn{1}{c|}{ACT} &20.8 &80.8 &17.8G &50.9M \\
&\multicolumn{1}{c|}{ARP (Ours)} &\textbf{24.8} &\textbf{94} &17.8G &\textbf{47.6M} \\\midrule
\multirow{3}{*}{RLBench} &\multicolumn{1}{c|}{RVT-2} &\multicolumn{2}{c|}{81.4} &57.1G &72.1M \\\cmidrule{2-6}
& \multirow{2}{*}{ARP (Ours)}  &\multicolumn{2}{c|}{81.6} &\textbf{56.2G} &\textbf{71.9M} \\
&  &\multicolumn{2}{c|}{\textbf{84.9}} &57.4G & 73.8M \\
\bottomrule
\end{tabular}}
    \caption{\textbf{Comparing our Autoregressive Policy to the SoTA of each environment.} Our autoregressive policy (ARP) outperforms environment-specific SoTA and is more efficient in MACs (number of multiply-accumulates) and parameter sizes.
    We report results of the transformer version of the diffusion policy because of its overall better performance. 
    \rbt{The RVT-2~\citep{goyal2024rvt2} results are reported from the original paper.}
    All MACs and parameter sizes are measured with THOP~\citep{thop}. We list the success rates of both the insertion (left)  and cube transfer (right) tasks in Gym-ALOHA~\citep{cadene2024lerobot}. 
    Per task success rates of RLBench are summarized in Figure~\ref{tab:rlbench-per-task}.}
    \label{fig:main-result}
    \vspace{-1em}
\end{figure}

\textbf{Inference.} During the test rollouts, we extract vision tokens from the current observation and provide them as input to ARP, which then generates actions autoregressively by sampling from the decoded action distribution and appending the selected actions to the existing action sequence.
This process of generating and executing actions is repeated until episode termination (success, failure, or reaching the time limit).
Actions are generated according to the sequence formats shown in Figure~\ref{fig:seq-design}. 
We manually set the chunk size for each type of action, and total sequence length for each task. We provide more implementation details and hyper-parameter values in Section~\ref{sec:experiment} and Appendix~\ref{ap:hp}.




\section{Experiments}
\label{sec:experiment}

In this section, we investigate how the Auto-regressive Policy (ARP) performs compared to the existing methods that were designed specifically for each environment.  
In addition, we examine whether auto-regression and action chunking are the primary contributors to the performance gains and evaluate how well existing methods perform across different environments.
Further, we verify ARP on a challenging nut-tightening task with a real robot. 
Finally, we demonstrate that ARP can estimate the likelihood of robot actions and predict actions based on human inputs. 
All of our source code and the pre-trained models can be found at \texttt{\href{http://github.com/mlzxy/arp}{http://github.com/mlzxy/arp}}. A single-file implementation of our ARP can be found at \texttt{arp.py}.

\subsection{Comparison with State-of-the-Art}


\textbf{Setup.} We compare the autoregressive policy (ARP) against the SoTA solutions in Push-T, ALOHA, and RLBench environments. Push-T is a single task. ALOHA consists of two tasks: insertion and cube transfer. RLBench includes 18 tasks, each with multiple language variants. These environments are illustrated in Figure~\ref{fig:env-intro} and Figure~\ref{fig:env-demo}. For Push-T and ALOHA, we train a separate policy for each task. For RLBench, a single policy is trained for all 18 tasks.
In Push-T, the policy observes the last two $96\times 96$ RGB frames, and predicts a window of future 2-d pointer positions as actions. In ALOHA, the policy observes the current $480\times 640$ RGB frame and predicts a window of future 14-dimensional joint positions. In RLBench, the policy observes four RGBD $128\times 128$ images and predicts the next target end-effector pose and gripper states. 
Existing SoTA techniques in these environments are outlined in Figure~\ref{fig:existing-sota}.
We use the same vision backbones as the SoTA solutions to extract vision tokens, namely ResNet50 
for Push-T and ALOHA, and Multi-View Transformer~\cite{goyal2024rvt2} for RLBench. 
We use the same training data, number of episodes, optimizer configuration, and evaluation frequency as the SoTA solutions. We detail the full list of hyper-parameters, such as the number of layers, sequence lengths, chunk sizes, and optimizer setups in Appendix~\ref{ap:hp}. 
Success rates for Push-T and RLBench are averaged over three independent runs. ALOHA's results are averaged over five runs.

\textbf{Results.} Figure~\ref{fig:main-result} shows that our autoregressive policy (ARP) \rbt{matches or} outperforms environment-specific SoTAs while being more computationally efficient. Figure~\ref{tab:rlbench-per-task} compares the per-task success rates of our ARP and RVT-2~\citep{goyal2024rvt2}. 
\rbt{We present two variants of ARP: both share the same autoregressive policy architecture, but the second variant has more CCT layers. The first ARP model matches the performance of RVT-2 while being more efficient, whereas the second one outperforms RVT-2. Notably, RVT-2 requires the current timestep as an input, whereas ARP relies solely on visual and language inputs. Further discussion on the architectural differences between RVT-2 and ARP and an analysis of the impact of timesteps, are provided in Appendix~\ref{apn:compare-with-rvt-2}.}


\subsection{Analysis}
\label{sec:analysis}

\begin{figure}\centering
\resizebox{\linewidth}{!}{
\begin{tabular}{cccc}\toprule
\multirow{2}{*}{Generation Mode} &\multirow{2}{*}{Push-T} &\multicolumn{2}{c}{ALOHA} \\\cmidrule{3-4}
& &Cube Transfer  &Insertion \\\midrule
SoTA & 78.8 & 80.8 & 20.8 \\
Action Chunking &77.6 &81.2 &21.2 \\
Single-token Autoregression &82.4 & 46 & 6.8 \\
Chunking Autoregression (Ours) &\textbf{87.1} &\textbf{94} &\textbf{24.8} \\
\bottomrule
\end{tabular}}
\caption{\textbf{Comparison of Action Prediction Strategies.} We compare the success rates (\%) of action prediction strategies shown in Figure~\ref{fig:comparison}. Action chunking and next-token autoregression are implemented by simply setting the chunk size to a constant of full sequence length or 1 in our ARP, while keeping other settings unchanged, such as action sequence design in Figure~\ref{fig:seq-design}. Our chunking autoregression supports variable chunk sizes and uses a different chunk size for each type of action token during generation.}\label{tab:generation-mode}
\vspace{-1em}
\end{figure}

\textbf{Does the performance gain come from chunking autoregression?} Our action sequence design incorporates additional inputs for Push-T and ALOHA, as shown in Figure~\ref{fig:seq-design}. These inputs are automatically extracted from the demonstration trajectories. In Push-T, the high-level waypoints are simply uniformly sampled down from the low-level trajectories and then discretized. In ALOHA, the pixel coordinates of the end-effector are computed from the joint values with the robot's forward kinematics and the camera parameters. It is possible that the performance gain of ARP originates from this extra information instead of our proposed architecture. 

Figure~\ref{tab:generation-mode} compares the success rates of action chunking, standard single-token autoregression, and our chunking autoregression in Push-T and ALOHA. 
They share the same implementation, with action chunking models generating the entire sequence at once by setting the CCT chunk size to the full sequence length, and single-token autoregression setting the chunk size to 1.
In contrast, our chunking autoregression uses different chunk sizes for each type of action. The results clearly show that our proposed chunking autoregression is the key factor behind the better performance.

Our approach innovates action chunking to support variable chunk sizes in sequence generation. Without our improvement, the standard next-token autoregression performs poorly at ALOHA, a task suites that requires fine-grained control inputs. We discuss why chunking autoregression matters in Fig.~\ref{fig:chunking-auto-regression} and Appendix~\ref{apn:discuss-act-chunk}. 
Compared to one-step action chunking models, our intuition can be explained through an example: imagine task $B$ is difficult, but solving task $A$ first, followed by solving task $B|A$ (task $B$ given the result of task $A$), is much easier. An autoregressive model follows this sequential process, solving task $A$ first and then leveraging the result to make task $B$ more feasible. In contrast, a one-step model attempts to predict both tasks simultaneously, treating $A$ and $B$ as independent problems. While the one-step model may solve task $A$ implicitly as part of solving task $B$, it does not explicitly take advantage of the problem structure and is therefore prone to shortcuts. This phenomenon has been explored in more depth for NLP tasks by \citet{prystawski2024stepbystep}. 


\begin{figure}\centering
\resizebox{\linewidth}{!}{
\begin{tabular}{lccccc}\toprule
\multirow{2}{*}{Method} &\multirow{2}{*}{PushT} &\multicolumn{2}{c}{ALOHA} &\multirow{2}{*}{RLBench} \\\cmidrule{3-4}
& &Cube Transfer & Insertion & \\\midrule
Diffusion Policy &78.8 &10 &1.6 &- \\
ACT &77.5 &80.8 &20.8 &69.8 \\
ARP (Ours) &\textbf{87.1} &\textbf{94} &\textbf{24.8} &\textbf{81.2} \\
\bottomrule
\end{tabular}}
\caption{\textbf{Evaluation of existing methods on various environments.} ACT, a VAE-based method, performs competitively across all environments, whereas Diffusion Policy struggles in ALOHA and RLBench. While we believe stronger diffusion-based methods can be developed in the future, our results suggest simpler architectures are more robust across diverse tasks.}\label{tab:cross-env}
\vspace{-1.5em}
\end{figure}

\textbf{Do existing methods work in different environments?} Figure~\ref{tab:cross-env} shows how existing methods perform in different environments. 
When testing in a new environment, we keep the same architecture but adapt the vision backbone and optimizer to the environment’s established setup.
RVT-2 was not implemented for Push-T and ALOHA, as it is designed for sparse waypoint predictions, which are incompatible with the high-frequency actions required in these tasks. We did not implement the diffusion policy for RLBench, as it refines actions from gaussian noise, which conflicts with the common practice in RLBench of predicting actions in discrete spaces. While 3D Diffuser Actor~\citep{ke20243ddiffactor} reports competitive results on RLBench, it uses a completely different architecture.

Figure~\ref{tab:cross-env} reveals that ACT, a VAE-based method, performs competitively across all environments, whereas the diffusion policy struggles to deliver meaningful performance in ALOHA. This outcome is surprising, given the recent popularity of diffusion-based techniques. While we believe a strong diffusion architecture, like 3D Diffuser Actor on RLBench, could be developed for ALOHA, this suggests that simpler architectures could be more robust across a wider range of tasks and environments. 
Our auto-regressive policy is trained with a single objective: to maximize the conditional likelihood of each action in a sequence. We believe this simplicity contributes to its robust performance across diverse environments.

\begin{figure}
    \centering
    \vspace{-1em}
    \includegraphics[width=0.7\linewidth]{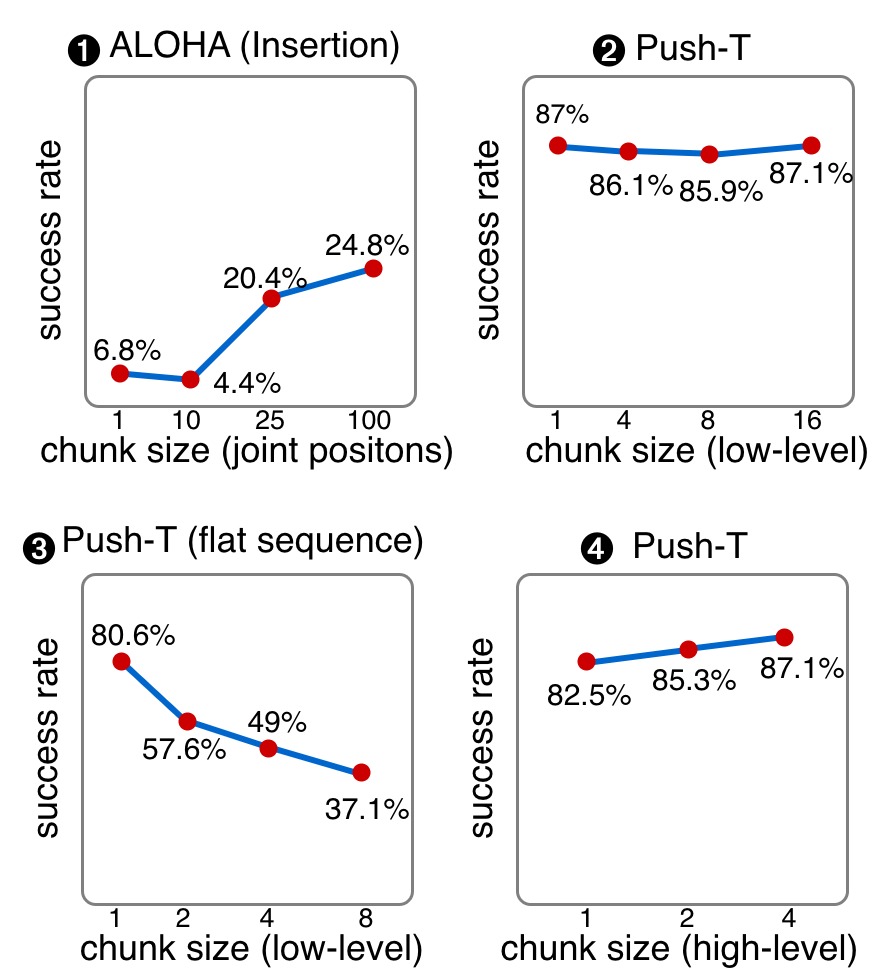}
    \vspace{-0.25em}
    \caption{\textbf{Impact of chunk size on performance.} Our results suggest that the optimal chunk size depends on both the task and the action sequence design. Therefore, the ability of our chunking causal transformer to support variable chunk sizes is essential for maximizing performance. }
    \label{fig:chunk-plot}
    \vspace{-1.5em}
\end{figure}

\textbf{How does chunk size affect performance?} Instead of predicting only the next token, our chunking causal transformer (CCT) is able to predict a variable number of next tokens, that is, a chunk of actions. Figure~\ref{fig:chunk-plot} illustrates the relationship between chunk size and success rate. The first plot shows that larger chunks significantly improve policy performance, a trend also observed by ACT~\citep{zhao2023aloha}. This advantage of chunking actions seems generalizable to high-frequency control in short-horizon tasks.
Interestingly, while larger chunk sizes for joint positions improve performance, action chunking without autoregression, where both end-effector waypoints and joint positions are predicted simultaneously, yields inferior results, as in Figure~\ref{tab:generation-mode}.

The second plot indicates that for Push-T, policy performance is largely insensitive to the chunk size of low-level trajectories because the standard deviation of the success rate ranges between 1 and 2. In this case, a moderate chunk size can be a better choice, given the common practice of executing only the first few predicted actions and then rerunning the policy, a test-time technique reduces error accumulation. This technique benefits from a moderate chunk size through early termination of autoregressive generation without sacrificing performance or computational efficiency.


In the third plot, we explore a different action sequence format for Push-T, where we remove high-level waypoints and flatten the trajectories into a vector, as detailed in Figure~\ref{fig:seq-design-pusht2}. 
This design yields a completely different trend, with the policy performing well only when the chunk size is 1.
The fourth plot shows that increasing the chunk size for high-level waypoints improves policy performance. 
These findings demonstrate that the optimal chunk size depends on both the task and the action sequence format.
As a result, CCT's ability to flexibly adjust variable chunk sizes is essential for maximizing performance.


\subsection{Real Robot Experiment}
\label{sec:real-robot}

\begin{figure}
    \centering
    \includegraphics[width=0.8\linewidth]{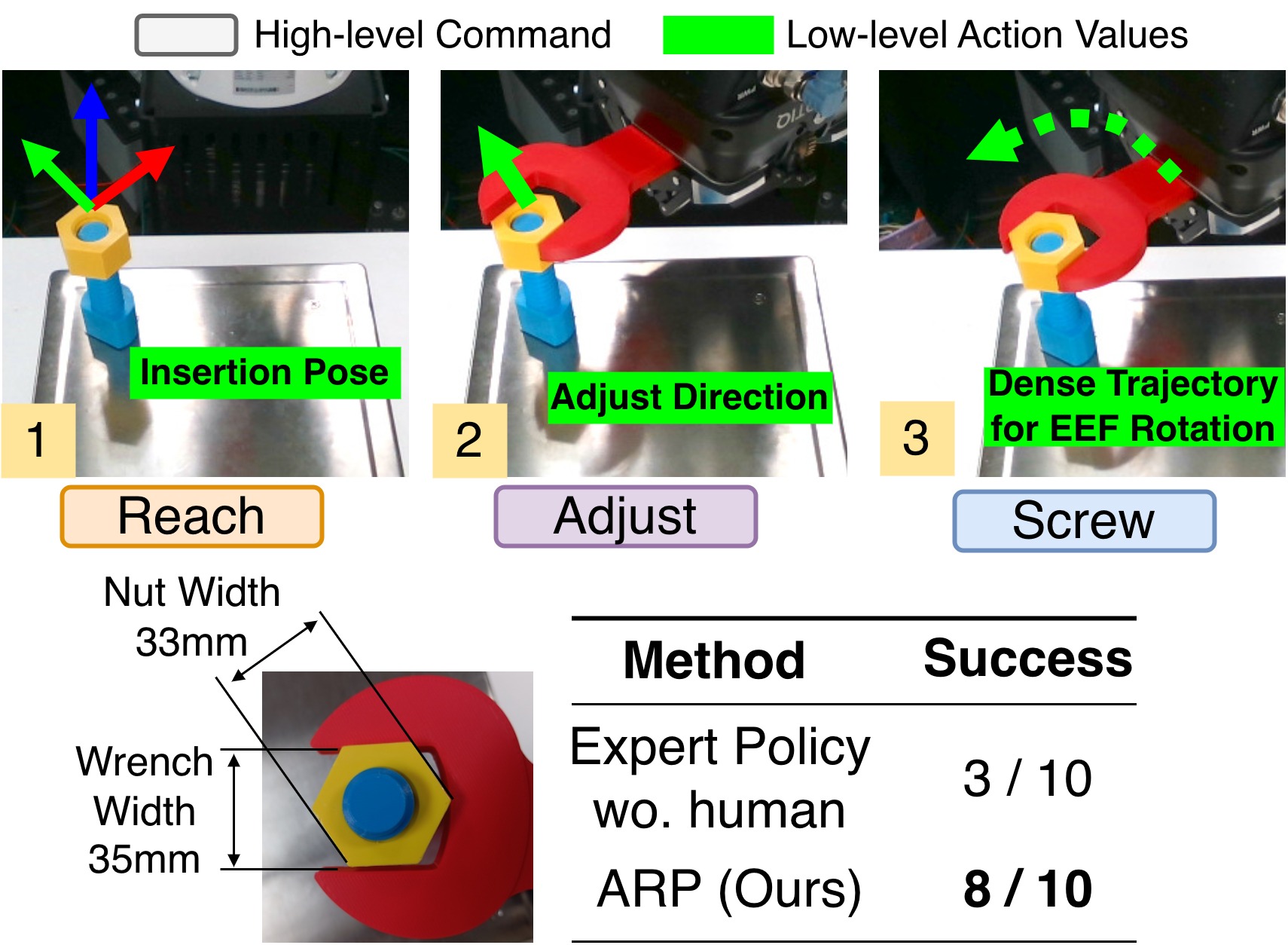}
    \caption{\textbf{Real robot experiment.} Our ARP learns to adaptively select high-level commands and generate low-level action values, including position adjustment after unsuccessful insertion. We achieve a success rate of 8/10 in this nut-screwing task that requires a precise tool alignment. The bolt's position (\textcolor{blue}{blue}) and nut's height (\textcolor{darkyellow}{yellow}) are randomized at every episode.}
    \label{fig:real-robot}
    \vspace{-1em}
\end{figure}

\textbf{Setup.} We evaluate ARP on a challenging tight nut-screwing task using a real robot, which requires precise alignment between the nut and wrench with a tolerance of 2mm, as shown in Figure~\ref{fig:real-robot}. 
In each episode, the bolt (\textcolor{blue}{blue}) is randomly placed on a 20$\times$20 cm$^2$ table, while the height of the nut (\textcolor{darkyellow}{yellow}) is randomized along the 6cm tall bolt. The orientations of both the bolt and nut are also randomized per episode. We define three primitive actions: reach, adjust, and bolt. At each step, our ARP predicts a high-level command to select the action and then generates corresponding low-level action values.
For example, ARP first predicts the reach command and an insertion pose. Next, the robot attempts to insert the wrench. 
After every unsuccessful attempt, the policy predicts the adjustment direction to adjust the wrench's position and reattempt insertion. 
Once the insertion succeeds, the policy switches to the screw command and predicts a dense trajectory to follow in order to rotate the end-effector around the wrench.
All commands are automatically predicted by the autonomous model instead of being manually specified. 
An impedance controller stops unsuccessful insertions based on force feedback. We deploy this model on a Kuka LBR iiwa robot. We use $480\times 640$ RBG-D observations from a single RealSense D415 camera. We use MVT as the vision backbone. 
To simplify the task, we assume the wrench is already grasped by the robot in a pre-defined position. An episode is considered successful if a screw action is completed after no more than three attempts to align the wrench on the nut. We trained ARP using 70 demonstrations collected from an expert policy. The expert policy uses Foundation Pose~\citep{wen2024foundationpose} to estimate insertion pose, with human operators providing fine-grained adjustments.

\textbf{Results.} Figure~\ref{fig:real-robot} shows that ARP screws nuts successfully in 8 out of 10 episodes, while the expert policy only has 3 successes out of 10 without human interventions. Most episodes succeeded without any adjustments because we used the adjusted successful insertion pose as the label for the reach command during training. To test ARP's adaptive adjustment ability, we add a uniform noise ranging from -5mm and 5mm along the normal plane of the insertion pose. Despite the noise, our ARP still succeeds in 6 out of 10 trials, with an average number of 1.6 adjustments per trial.

\subsection{Qualitative Visualization}

We showcase all the evaluation tasks in Figure~\ref{fig:env-demo}. Video demonstrations of ARP in simulation and in the real world are available in the supplementary material. In this section, we show two key advantages of ARP: (1) estimating the likelihood of given robot actions, (2) and predicting actions conditioned on human input. 

\begin{figure}
    \centering
   
    \includegraphics[width=\linewidth]{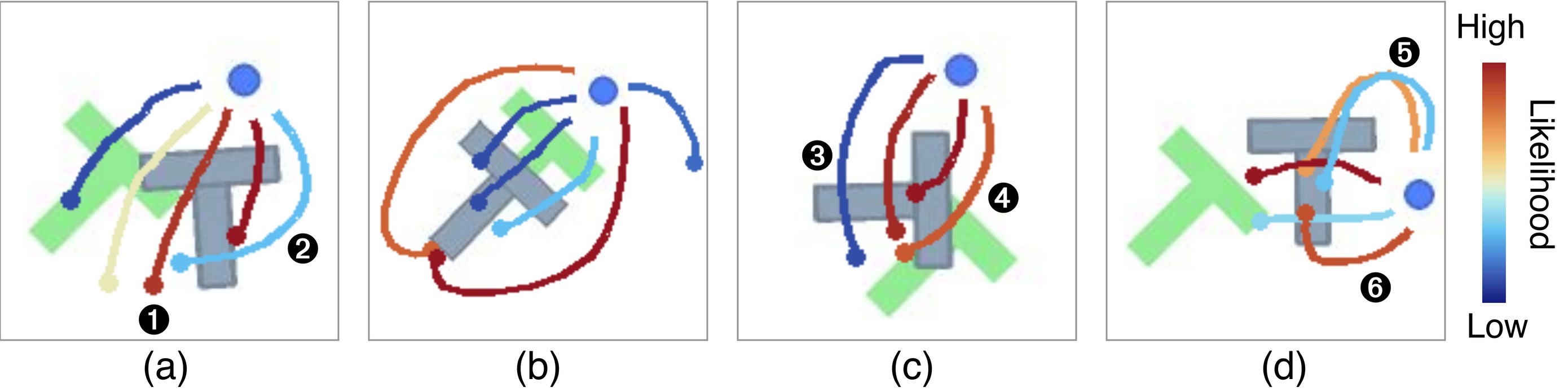}
     \vspace{-1.5em}
    \caption{\textbf{Trajectory Likelihood Estimation.} ARP generally assigns higher likelihoods to effective trajectories over futile ones, and demonstrates its understanding of action multi-modality as in subfigure (b). ARP's likelihood inference ability can identify model weaknesses and find defective demonstrations. All trajectories are human-drawn and are not seen in training.} 
    \label{fig:likelihood}
    \vspace{-1em}
\end{figure}

\begin{figure}
    \vspace{-0.5em}
    \centering
    \includegraphics[width=\linewidth]{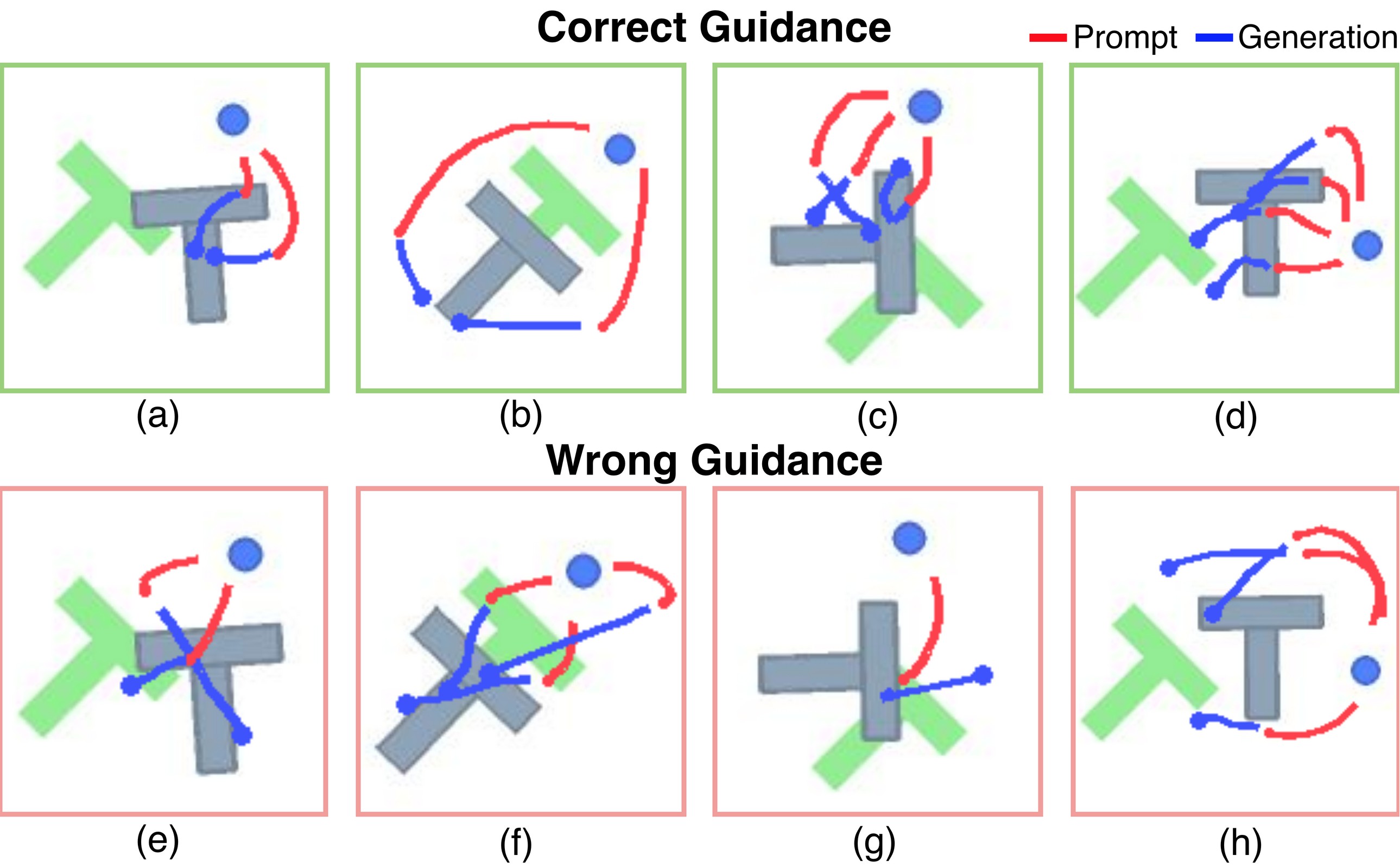}
    \vspace{-1.5em}
    \caption{\textbf{Trajectory Prediction based on Human Guidance.} We show predicted trajectories of ARP (\textcolor{blue}{blue}), conditioned on human-drawn trajectories (\textcolor{red}{red}). The correct guidance is given with the intention of completing the task, and the wrong guidance is aimed at failing the task. ARP performs as expected under correct guidance. Under wrong guidance, ARP recovers from failure in subfigure (g), avoids further mistakes in subfigure (h), and amplifies the errors in subfigure (e) and (f), which reflects out-of-distribution behavior, as the training set consists only of successful demonstrations.} 
    \label{fig:human-guide}
    \vspace{-1em}
\end{figure}

\textbf{Likelihood inference.} To generate the next token $a_n$, an auto-regressive model estimates the conditional probability $P(a_n|a_1,...,a_{n-1})$. Using the product rule, the model can estimate the joint probability $P(a_1, ..., a_n) = \prod_{i=2}^{n} P(a_i | a_1, \dots, a_{i-1}) P(a_1)$ for any given sequences, a capability that more advanced generative frameworks such as VAE and diffusion lack. Figure~\ref{fig:likelihood} shows  for different trajectories the likelihood estimated by ARP. All these trajectories are human demonstrations. ARP generally assigns higher likelihoods to effective trajectories and lower likelihoods to futile ones. For instance, in sub-figure (b), ARP assigns high likelihoods to two symmetrical trajectories around the T object, demonstrating its understanding of action multi-modality. However, some likelihood assignments are less intuitive. For example, trajectories \ding{202}, \ding{205}, and \ding{207} receive moderately high likelihoods, yet they may not bring the T-shape object closer to the green target, at least not better than the low-likelihood trajectories \ding{203} and \ding{204}. \ding{206} marks two similar trajectories, yet they have different likelihoods. We believe that this type of likelihood inference can help identify the model's weaknesses and eliminate defective demonstrations from the training data.

\textbf{Prediction with human guidance.} Auto-regressive models generate future tokens conditioned on the previous sequence. In Figure~\ref{fig:human-guide}, we illustrate examples of trajectories of ARP (\textcolor{blue}{blue}) in Push-T, predicted conditionally on human-drawn initial trajectories (\textcolor{red}{red}). The first row (\textcolor{darkgreen}{green}) shows predictions under correct guidance, where the intention is to complete the task successfully. The second row (\textcolor{pink}{pink}) is based on a wrong guidance with the intention of failing the task. ARP completes the trajectory correctly given a correct initial part. Given a wrong initiation, sub-figure (g) shows ARP's recovery from failure by correcting its initial trajectory. In sub-figures (e) and (f), however, ARP amplifies the initial error by pushing further in the wrong direction. This behavior reflects ARP’s out-of-distribution response, as the training set consists only of successful demonstrations.

\section{Discussion}
\label{sec:discussion}


We have shown that ARP is a strong and universal architecture that can be trained to perform diverse manipulation tasks. Here we discuss its limitations and potential future directions.

\textbf{Learning to plan.} 
Planning is a key ability of intelligent agents.
It requires the agent to reason not only about its actions but also their impacts on its environment. 
Motivated by the reasoning capacity of auto-regressive models in NLP, a promising direction is to incorporate planning into ARP. 
One possible solution is to predict sequences of both states and actions. 
States in robotics are typically high-dimensional, such as images or point clouds. 
To solve this problem, ARP can be extended to generate future states by using recent hybrid architectures of autoregression~\citep{chen2024diffusionforcing}.

\textbf{Interactive robot learning.} Human-Robot collaboration improves efficiency by allowing the robot to recover from its errors~\citep{liu2023interactive}. One possible future direction is to integrate active learning techniques into ARP  to learn from immediate human feedback. The auto-regressive mechanism naturally supports conditioning action prediction on human input. 
Moreover, ARP can estimate the likelihood of action sequences. Likelihood is a common measure for identifying the most informative samples in active learning~\citep{taylor2021active}. This can be used, for example, to prioritize demonstrations of tasks where the robot encounters more difficulties.

\textbf{Adaptive action sequence learning.} Despite ARP's impressive performance, it still requires a manual design of action sequence formats and chunk sizes for each environment.
Unlike natural language, robot actions lack a universal vocabulary. 
A promising direction is to design a universal robot action language applicable across various environments~\cite{zhang2024vkt, zheng2025universal}, which reduces the cost of defining new actions, unifies training datasets, and improves generalization.

\bibliographystyle{IEEEtran}


\clearpage
\appendix
\section{Appendix}
\label{sec:appendix}
\renewcommand{\thefigure}{A\arabic{figure}}
\renewcommand{\theHfigure}{A\arabic{figure}}
\setcounter{figure}{0}
\renewcommand{\thetable}{A\arabic{table}}
\renewcommand{\theHtable}{A\arabic{table}}
\setcounter{table}{0}

\subsection{Code and Pretrained Models}

The source code of our autoregressive policy is included in both the supplementary folder \texttt{Code} and \url{https://github.com/mlzxy/arp}. Please check the \texttt{README.md} for instructions on installation, dataset setup, downloading pretrained models, and documentation.

\subsection{Hyper-parameters and Implementation Details}
\label{ap:hp}

In this section, we provide a full list of hyper-parameters in Figure~\ref{tab:hp-pusht}, Figure~\ref{tab:hp-aloha}, and Figure~\ref{tab:hp-rlb} for Push-T, ALOHA, and RLBench, respectively, along with comments on selected hyper-parameters to provide additional implementation details. 

\textbf{Model.} The mlp size denotes the hidden feature dimension of the MLP network within the standard multi-head attention operation. The number of latents refers to the number of Gaussians for the Gaussian mixture distribution used to decode continuous actions. The backbone denotes the network used to extract the vision features. We use the ResNet50 for Push-T and ALOHA, and Multi-View Transformer (MVT) for RLBench, identical to the ones used in Diffusion Policy, ACT, and RVT2.

\textbf{Action Sequence.} The horizon refers to the number of actions predicted at each step, while the number of action steps indicates how many actions are actually executed, with the remainder discarded. We adopt the same horizon and action steps as state-of-the-art methods. In Push-T, the chunk size for both high- and low-level actions matches the horizon, meaning all high-level points are predicted in one chunk, followed by all low-level points. Yet, interestingly, as shown in Figure~\ref{tab:generation-mode}, combining these two chunks into a single-step prediction degrades performance. For RLBench, which uses the next key end-effector pose as the control interface, there is no need for high-frequency actions, so neither the horizon nor action steps apply. Instead, low-level robot movements are generated using RLBench's built-in RRT planner. We use a chunk size of 2 for binary gripper states and a chunk size of 1 for end-effector positions and rotations. For example, ARP first predicts the roll, followed by pitch and yaw of the rotation Euler angle.  We follow the strategy of RVT-2 to predict coarse positions and then refine them by zooming into the images (with updated vision features) to obtain more accurate positions. The end-effector positions are predicted in 2-d, and the 3-d positions are derived from the 2-d coordinates of each viewpoint.

\begin{figure*}
    \centering
    \includegraphics[width=\linewidth]{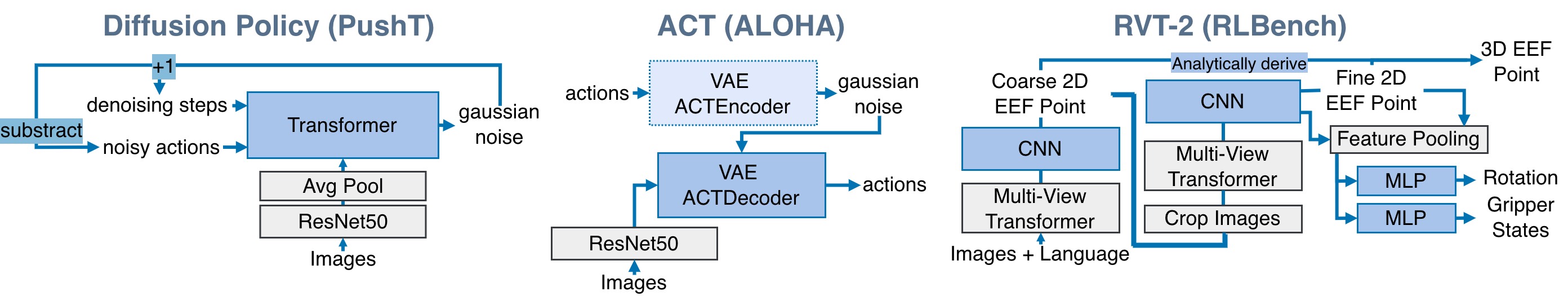}
    \vspace{-1em}
    \caption{\textbf{Overview of SoTA solutions on Push-T, ALOHA, and RLBench.} Diffusion Policy (DP)~\citep{chi2023diffp} iteratively subtracts Gaussian noises from noisy actions. The transformer network predicts the Guassian noise at each step.
    Action Chunking Transformer (ACT)~\citep{zhao2023aloha} is a VAE architecture that predicts actions directly from images and Gaussian noises. 
    RVT-2~\citep{goyal2024rvt2} is a hybrid and more complex model, but it is trained directly with behavior cloning and it 
    does not require a generative framework such as diffusion or VAE. }
    \label{fig:existing-sota}
\end{figure*}

\textbf{Train \& Eval.} The observation $2 \times 96 \times 96 \times 3$ represents 2 frames of RGB images, each with a resolution of 96x96 pixels. For RLBench, the observation \(4 \times 128 \times 128 \times 4\) refers to RGBD images (with one depth channel) at 128x128 resolution, captured from 4 cameras.
In ALOHA, the maximum evaluation steps of 400 and control frequency of 50Hz indicate an evaluation time limit of 8 seconds. LAMB refers to the large batch optimizer. We use the same number of training steps, evaluation frequency, optimizer, learning rate, and learning rate scheduler as used by the SoTA solutions.

\begin{table}[h]\centering
\caption{Hyperparameters used in our experiments on Push-T.}\label{tab:hp-pusht}
\begin{tabular}{lr}\toprule
\textbf{Hyperparameter} &\textbf{Value} \\\midrule
\multicolumn{2}{l}{\textit{Model}} \\\midrule
number of layers &30 \\
embedding size &64 \\
mlp size &256 \\
number of latents (gmm) &4 \\
backbone &RN50 \\\midrule
\textit{Action Sequence} & \\\midrule
horizon (low-level) &16 \\
horizon (high-level) &4 \\
number of action steps &8 \\
chunk size (low-level) &16 \\
chunk size (high-level) &4 \\\midrule
\textit{Train \& Eval} & \\\midrule
observation &RGB $2\times 96\times 96 \times 3$ \\
control frequency &10 \\
maximum evaluation steps &300 \\
train epochs &2000 \\
eval frequency &50 \\
batch size &128 \\
learning rate &0.0001 \\
learning rate scheduler &cosine with restart \\
optimizer &AdamW \\
\bottomrule
\end{tabular}
\end{table}

\begin{table}\centering
\caption{Hyperparameters used in our experiments on ALOHA}\label{tab:hp-aloha}
\begin{tabular}{lr}\toprule
\textbf{Hyperparameter} &\textbf{Value} \\\midrule
\multicolumn{2}{l}{\textit{Model}} \\\midrule
number of layers &4 \\
embedding size &512 \\
mlp size &2048 \\
number of latents (gmm) &1 \\
backbone &RN50 \\\midrule
\multicolumn{2}{l}{\textit{Action Sequence}} \\\midrule
horizon (joints) &100 \\
horizon (waypoints) &10 \\
number of action steps &100 \\
chunk size (joints) &100 \\
chunk size (waypoints) &1 \\\midrule
\textit{Train \& Eval} & \\\midrule
observation &RGB $1\times 480 \times 640 \times 3$ \\
control frequency &50 \\
maximum evaluation steps &400 \\
train steps &100000 \\
eval frequency &10000 \\
batch size &8 \\
learning rate &1.00e-5 \\
learning rate scheduler &none \\
optimizer &AdamW \\
\bottomrule
\end{tabular}
\end{table}

\begin{table}\centering
\caption{Hyperparameters used in our experiments on RLBench.}\label{tab:hp-rlb}
\begin{tabular}{lr}\toprule
\textbf{Hyperparameter} &\textbf{Value} \\\midrule
\multicolumn{2}{l}{\textit{Model}} \\\midrule
number of layers &8 \\
embedding size &128 \\
mlp size &512 \\
backbone &MVT \\\midrule
\textit{Action Sequence} & \\\midrule
chunk size &mix of 2 and 1 \\\midrule
\textit{Train \& Eval} & \\\midrule
observation &RGBD $4\times 128\times 128 \times 4$ \\
maximum evaluation steps &25 \\
train epochs &80000 \\
eval frequency &10000 \\
batch size &96 \\
learning rate &1.25e-5 \\
learning rate scheduler &cosine \\
optimizer &LAMB \\
\bottomrule
\end{tabular}
\end{table}

 \begin{figure}[h!]
    \centering
    \includegraphics[width=0.8\linewidth]{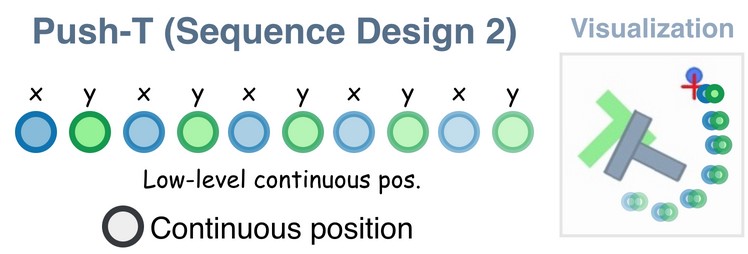}
    \caption{\textbf{Flattened Action Sequence for Push-T.} Based on the action sequence in Figure~\ref{fig:seq-design}, we remove the high-level waypoints and flatten the 2D coordinates into a single vector. For example, a trajectory of $(x_1, y_1), (x_2, y_2), (x_3, y_3)$ is transformed into vector $(x_1, y_1, x_2, y_2, x_3, y_3)$. The policy is trained to predict first the x-coordinate of the initial point, then the y-coordinate, followed by the x- and y-coordinates of subsequent points.} 
    \label{fig:seq-design-pusht2}
\end{figure}

\subsection{Discussion on Action Chunking}
\label{apn:discuss-act-chunk}

Action chunking has a clear downside -- when predicting multiple actions at a time, the agent doesn't receive information about what state was observed after the first action. This means that the agent is operating with less information than if a single-step prediction was used. At the same time, in a MDP the state is guaranteed to be a sufficient statistic for the optimal policy. Given this information, why should action chucking be useful?

\begin{figure}
    \centering
    \includegraphics[width=\linewidth]{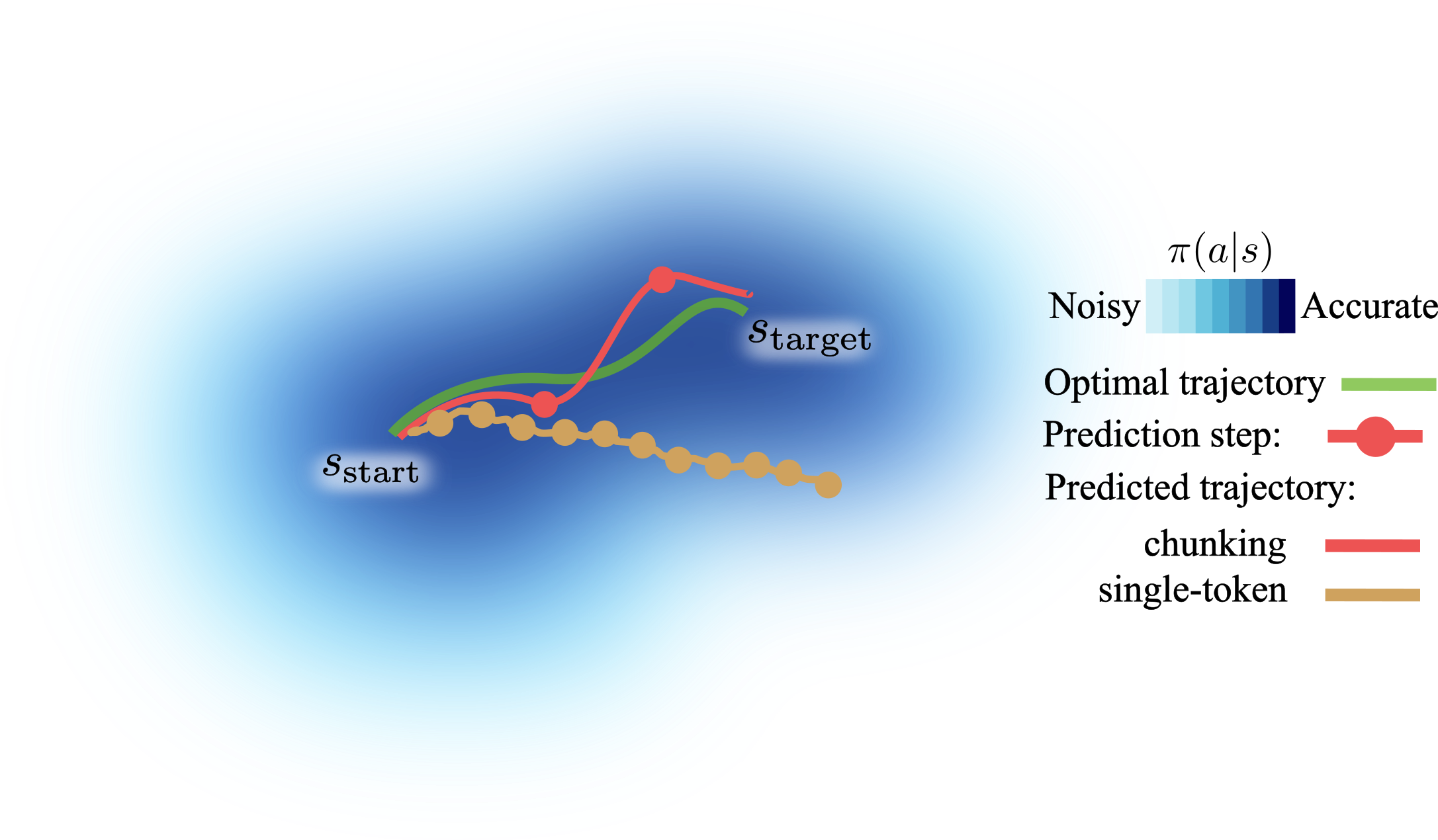}
    \caption{\textbf{Why Chunking Autoregression Works.} Consider a robot navigating from state $s_{\text{state}}$ to $s_{\text{target}}$ in configuration space, where policy accuracy is indicated by color intensity (darker = higher accuracy). The green line denotes the optimal trajectory. Chunking autoregression has smaller accumulated error by having fewer prediction steps, keeping the trajectory within high-accuracy regions and converging near $s_{\text{target}}$. In contrast, single-step autoregression suffers from error compounding---each step introduces noise, progressively pushing predictions into out-of-distribution regions. Crucially, this divergence occurs regardless execution, as autoregressive generation conditions on previous predictions: noisy input from last step lead to increasingly unstable outputs.}
    \label{fig:chunking-auto-regression}
\end{figure}

We propose two main reasons. First, as has been explored in other imitation learning works, using expert data means that the dataset often lacks information on how to recover from errors, which means that predictions grow worse over time. Using longer action chunks effectively shortens the time horizon. However, we find that action chunking still has noticeable benefits even when the state is well-covered, such as in the Push-T environment. Additionally, this problem becomes less severe as the dataset grows -- when the prediction error goes to zero, so does the effect of error recovery.

\begin{figure}[h!]
    \centering
    \includegraphics[width=\linewidth]{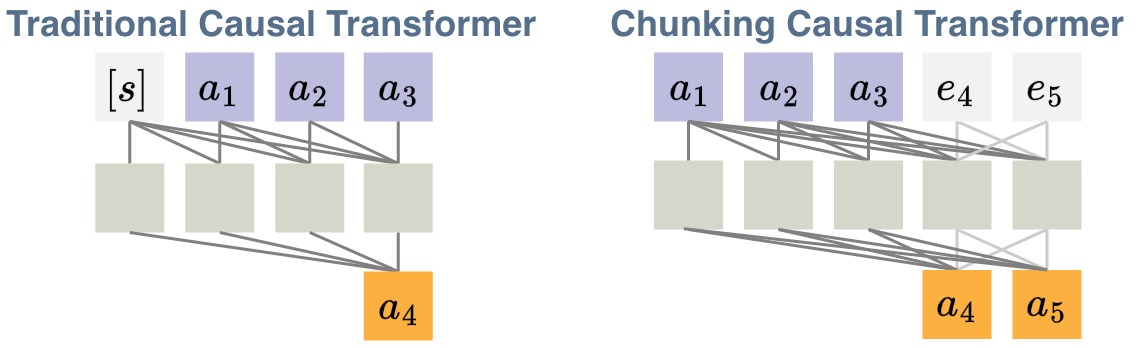}
    \vspace{-1em}
    \caption{\textbf{Causal Transformer versus Chunking Causal Transformer.} Causal transformer prepends the input sequence with a \enquote{start} token $[s]$ and modifies the token embedding with causal attention so that the last token $a_3$ becomes the next token $a_4$. Chunking Causal Transformer (CCT) appends the input sequence with a chunk of empty tokens, for example, $e_4, e_5$. CCT modifies the token embedding with causal attention for the action tokens $a_1, a_2, a_3$ and bidirectional attention for the empty tokens $e_4, e_5$. The empty tokens $e_4, e_5$ become the next tokens $a_4, a_5$. CCT can predict a variable amount of next tokens by configuring the number of empty tokens. }
    \label{fig:compare-cct-ct}
\end{figure}

The second and perhaps stronger explanation is that if the demonstrations are non-Markov, the Markov policy that maximizes single-step accuracy is *not necessarily the optimal policy*. This is true even even if the demonstration policies are optimal, and even in the limit as data and model capacity become infinite. This is because the state occupancy measure is not convex with respect to the policy, so linear combinations of policies can lead to state distributions that are not linear combinations of the demonstration state distributions. 
This can be address either by learning a non-Markov policy, or by learning a Markov policy that imitates the desired state distribution rather than the demonstrations.

\begin{figure*}
    \centering
    \includegraphics[width=\ral{0.8}\linewidth]{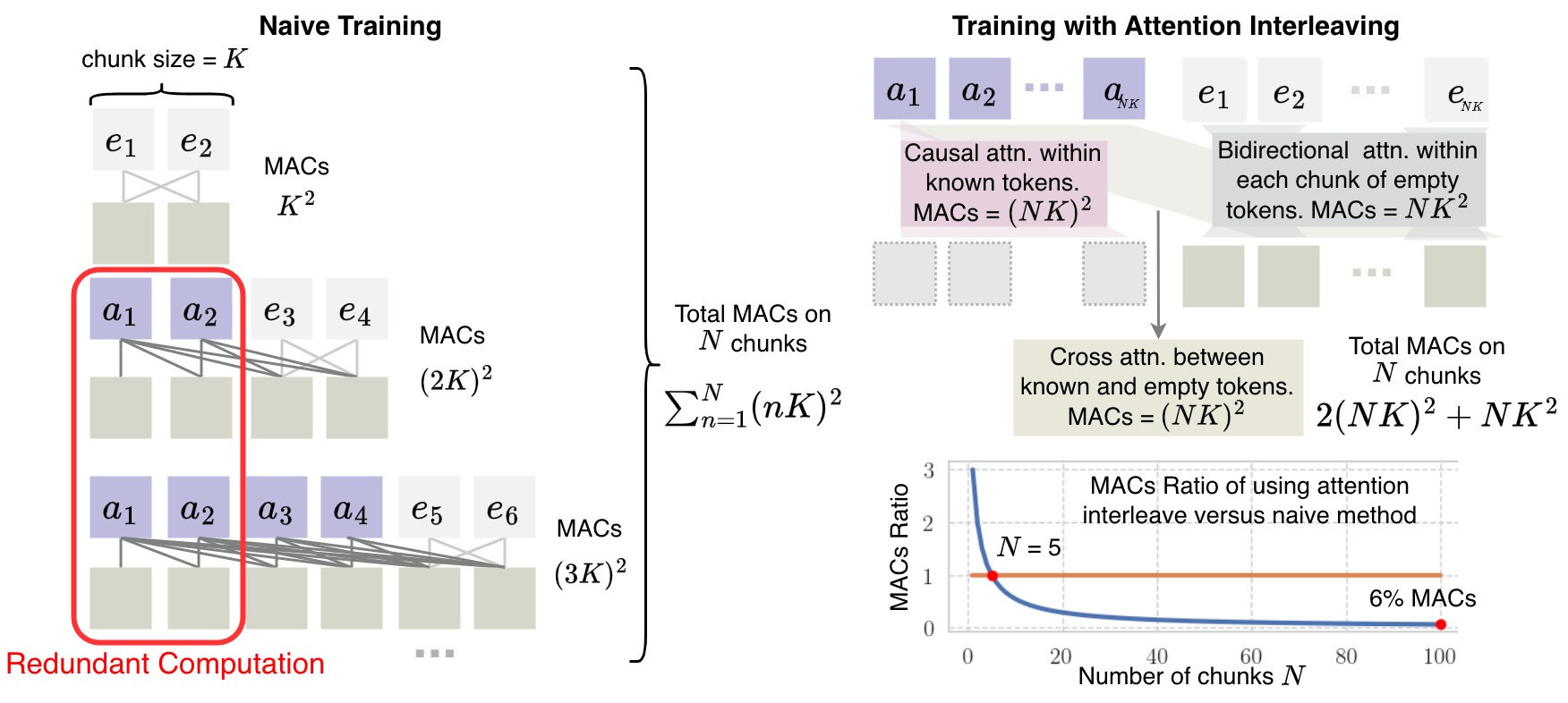}
    \caption{\textbf{Naive Training versus Training with Attention Interleaving.} The left figure demonstrates 
 that the causal attention within $a_1, a_2$ is computed twice, when inputs are $a_1, a_2, e_3, e_4$ and  $a_1, a_2, a_3, a_4, e_5, e_6$. This redundancy can be reduced by precomputing the causal attention of all action tokens and caching the results. 
    In doing so, the MACs are reduced from $\sum_{n=1}^N(nK)^2$ to $2(NK)^2 + NK^2$, where $N$, $K$ are chunk number and chunk size.
    For simplicity, we count the MACs as the number of attention entries. In addition to the reduced MACs, we find that having a single forward pass for all tokens yields a much cleaner training procedure, a benefit that is not quantified by the raw number of multiply-accumulate operations. }
    \label{fig:attention-interleave}
\end{figure*}

\begin{figure*}[!htp]\centering
\resizebox{\linewidth}{!}{
\begin{tabular}{lcccccccccc}\toprule
\multirow{2}{*}{Method} &Avg. &Avg. &Close &Drag &Insert &Meat off &Open &Place &Place &Push \\\cmidrule{2-11}
&Success &Rank &Jar &Stick &Peg &Grill &Drawer &Cups &Wine &Buttons \\\midrule
RVT2 &81.4 &2.22 &\textbf{100.0} &99.0 &40.0 &\textbf{99.0} &74.0 &38.0 &95.0 &\textbf{100} \\
ARP (Ours) &81.6 &1.89 & 97.6 &88.0 &53.2 &96.0 &90.4 &48 &92.0 &\textbf{100.0} \\
ARP$^+$ (Ours) &\textbf{84.9} &\textbf{1.61} &95.2 &\textbf{99.2} & \textbf{78.4} & 97.6 &\textbf{92.8} & \textbf{48.8} & \textbf{96} &\textbf{100.0} \\\midrule
&Put in &Put in &Put in &Screw &Slide &Sort &Stack &Stack &Sweep to &Turn \\
&Cupboard &Drawer &Safe & Bulb & Block & Shape & Blocks & Cups & Dustpan & Tap \\\midrule
RVT2 &66.0 &96.0 &\textbf{96.0} & 88.0 &92.0 &35.0 &\textbf{80.0} &69.0 &\textbf{100.0} &99.0 \\
ARP (Ours) &68.0 &\textbf{99.2} &94.4 &85.6 & \textbf{98.4} & 35.2 & 55.2 &76.8 &90.4 &\textbf{100.0} \\
ARP$^+$ (Ours) & \textbf{69.6} &98.4 &86.4 &\textbf{89.6} &92.8 & \textbf{46.4} &63.2 &\textbf{80.0} &97.6 &96.0 \\
\bottomrule
\end{tabular}}
\caption{\textbf{Performance on RLBench.} We report the success rate for each task, and measure the average success rate and rank across all tasks. ARP$^+$ shares the same network definition with ARP but has more layers. The MACs / parameter sizes of RVT-2, ARP, ARP$^+$ are 72.1M/57.1G, 71.9M/56.2G, and 73.8M/57.4G, respectively. ARP performs comparably or outperforms RVT-2 on all tasks. \rbt{Note that RVT-2 requires current timestep as input, and ARP models do not use timestep.}}\label{tab:rlbench-per-task}
\end{figure*}

\begin{figure*}
    \centering
    \includegraphics[width=\arxiv{0.8}\ral{0.7}\linewidth]{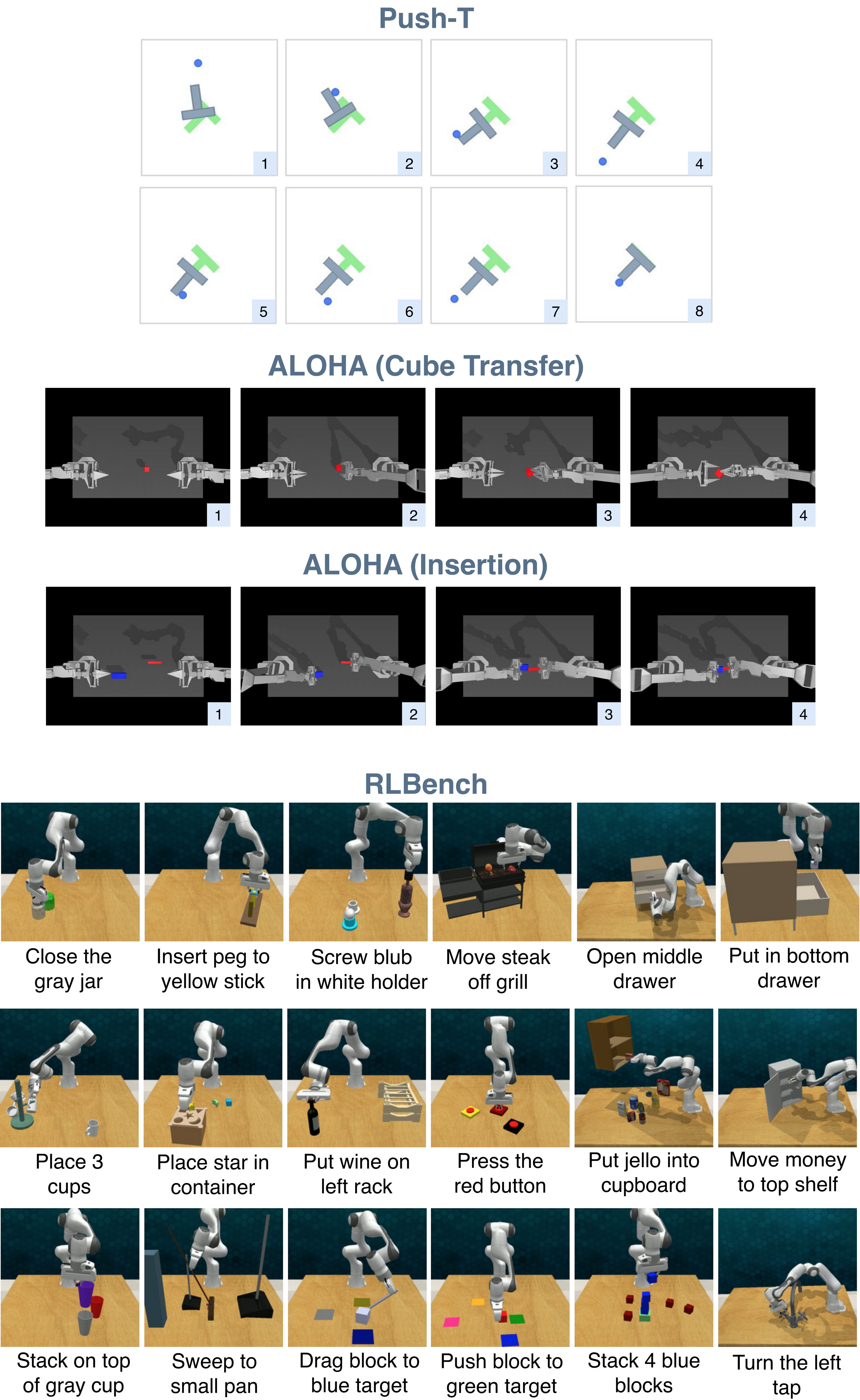}
    \caption{\textbf{Demonstrations of all tasks in Push-T, ALOHA, and RLBench.} We provide visualizations of key frames from a single episode of Push-T and ALOHA, with the frame order indicated at the bottom right. For RLBench, we visualize one language variant for each task. RLBench features over 100 task variants specified through natural language commands~\citep{james2020rlbench}, such as \texttt{"open [pos] drawer"} where \texttt{pos} is selected from \texttt{top, middle, bottom}, and \texttt{"stack [num] [color] blocks"}, where \texttt{num} ranges from \texttt{2, 3, 4}, and \texttt{color} is chosen from a palette of 20 colors.}
    \label{fig:env-demo}
\end{figure*}

\rbt{\subsection{Comparison with RVT-2}}
\label{apn:compare-with-rvt-2}

\noindent\rbt{\textbf{Architectural difference between RVT-1 and ARP.}} Our ARP shares the same visual-language encoder, i.e., Multi-View Transformer (MVT) with RVT-2. Three notable architectural differences between ARP and RVT-2 are: 

\begin{enumerate}
    \item RVT-2 employs a two-stage approach. In the first stage, RVT-2 predicts a coarse end-effector pose from images of the entire workspace. New images are then captured at the predicted coarse pose location, providing finer visual details of the surrounding area. In the second stage, RVT-2 uses these detailed images to predict a fine-grained end-effector pose, which serves as the final output.  To implement this, RVT-2 utilizes two MVT encoders---one for coarse inputs and another for fine inputs---along with two separate policy networks for coarse and fine predictions. Each policy network is composed of CNNs and MLPs.
    
    Similarly, our ARP adopts this two-stage approach for RLBench tasks. We also employ two MVT encoders for coarse and fine inputs, respectively. However, unlike RVT-2, ARP uses a single autoregressive policy network. During the coarse stage, the input visual tokens to this network are from the coarse MVT encoder, while during the fine stage, the input visual tokens come from the fine MVT encoder.

    \item  RVT-2 uses \enquote{Location Conditioned Rotation}, a handcrafted MLP that predicts gripper rotation conditioned on the gripper's translation. In contrast, ARP achieves a similar effect through autoregression, eliminating the need for a manually designed component.

    \item  RVT-2 directly upscales the visual features from the MVT encoder to predict pixel coordinates. In contrast, ARP upscales the multiplication of the predicted token (generated through autoregression) with the MVT visual features. Our approach is more aligned with sequence learning, as shown in Figure 7.

\end{enumerate}

We remark that all three architectural differences pertain specifically to the autoregressive policy and are not part of the visual-language encoder.

\vspace{1em}\noindent\rbt{\textbf{Why not use timestep as model input.}} In Figure~\ref{fig:main-result}, the RVT-2 result is obtained with the current timestep as input, while our ARP models do not include timestep in their input.
We do not use timestep as input for two reasons. First, during training, RVT-2 does not utilize the correct timestep due to an implementation bug in its data loader. This issue can be traced down in the RVT-2 codebase at 
\href{https://github.com/NVlabs/RVT/blob/faf459d16ec6c18bf43f8f0c55d372d73dde076a/rvt/utils/dataset.py#L239}{L239}, \href{https://github.com/NVlabs/RVT/blob/faf459d16ec6c18bf43f8f0c55d372d73dde076a/rvt/utils/dataset.py#L265}{L265}, \href{https://github.com/NVlabs/RVT/blob/faf459d16ec6c18bf43f8f0c55d372d73dde076a/rvt/utils/dataset.py#L392}{L392} of \texttt{rvt/utils/dataset.py}. Specifically, the timestep of a frame varies depending on when it is inserted into the replay buffer, which we believe is an unintended behavior by the authors of RVT-2. Accurately emulating this behavior, or migrating our implementation to use this data loader would require substantial engineering effort. Moreover, this data-loader has been noted by other researchers for being confusing, as highlighted in this GitHub discussion (\href{https://github.com/peract/peract/issues/6}{link}) and even the code comments from RVT (\href{https://github.com/NVlabs/RVT/blob/faf459d16ec6c18bf43f8f0c55d372d73dde076a/rvt/utils/dataset.py\#L240}{link}). Therefore, we have opted to train ARP without using time-step information. 

Second, and more importantly, the timestep in RLBench does not correspond to physical time but rather to the number of macro-steps executed so far in the current episode. Each macro-step consists of three stages: (1) predicting the next end-effector pose and gripper action using a policy (e.g., RVT-2 or ARP), (2) planning a trajectory to reach the predicted pose using RRT, and (3) executing the trajectory and gripper action. As noted by the authors of RVT-2 and PerAct in these GitHub discussions (\href{https://github.com/NVlabs/RVT/issues/22}{link1}, \href{https://github.com/peract/peract/issues/31}{link2}), \textquote{time could matter as it informs the network of the current stage of the task}. However, in the real world, the autonomous robot must learn to infer the task stage solely from visual inputs, as there will be no oracle that will be telling the robot the current stage of the task. This capability is critical for scenarios where the robot must interact with the physical world continuously without requiring controlled resets to an initial stage with a timestep of 0, such as placing objects in predefined regions. For these reasons, we have chosen to exclude timestep information when training ARP models.

\vspace{1em}\noindent\rbt{\textbf{Dive deeper into timestep and sampling strategy.}} As we mention above, existing methods for RLBench, such as PerAct, RVT, and ARM are impacted by a flawed training data-loader implementation. The behavior of this data-loader is two-fold: 
\begin{enumerate}
    \item The sampling rate of keyframes gets increased. Note that keyframes are the frames where the gripper stopped, which are provided by RLBench.
    \item The timestep of a frame depends on when it is inserted into the replay buffer (randomized).
\end{enumerate}

As a result of its unclear behavior, it is difficult to study the impact of timestep or sampling strategy in training RLBench models. To resolve this issue, we implemented our RLBench dataloader from scratch. Our dataloader has a simple implementation and straightforward behaviors: 

\begin{enumerate}
\item It only loads keyframes.
\item It provides a correct, non-randomized timestep, if timestep is required as input. 
\end{enumerate}

Table~\ref{tab:abla-timestep} compares the results of RVT-2 and ARP on different frame sampling strategies and timestep configurations. The \enquote{original sampling + randomized timestep} represents the official implementation of RVT-2. We have some surprising observations: 

\begin{enumerate}
    \item If we remove the randomized timestep, then the performance of RVT-2 drops significantly (81.4 to 77.0). This indicates that the timesteps, due to the unintended randomization, may serve as a regularization instead of providing extra information.

    \item 
    \finalrev{The original implementation of RVT-2 requires the timestep as an input. However, we find that by
updating the dataloader, we can achieve comparable results without it (81.6 vs. 81.4).} This indicates the importance of keyframes in RLBench training, and also verifies the effectiveness of our implementation. 

    \item If correct timesteps are provided during training, the performance of both RVT-2 and ARP drops drastically (81.4 to 74.1, 81.6 to 77.8). This indicates that the correct timestep is actually harmful to RLBench model trainings, contrary to previous believes.  This can be explained by seeing timestep as an information leakage for task stage. 
\end{enumerate}

\begin{table}[h]\centering
\caption{Impacts of sampling strategy and timestep on RLBench models. The original data-loader of RVT-2 is adopted from prior works, including RVT, PerAct, and ARM. However, this implementation has been noted by researchers as being confusing (\href{https://github.com/peract/peract/issues/6}{discussion}, \href{https://github.com/NVlabs/RVT/blob/faf459d16ec6c18bf43f8f0c55d372d73dde076a/rvt/utils/dataset.py\#L240}{comment}). Moreover, as detailed in Appendix~\ref{apn:compare-with-rvt-2}, the original data-loader randomizes the timestep of training samples, an unintended behavior by the previous authors. In contrast, we propose a simplified yet equally effective approach that only samples keyframes, which enables the study of the impacts of timestep and provides a solid foundation for future research.}\label{tab:abla-timestep}
\resizebox{\linewidth}{!}{
\begin{tabular}{c|c|c|c}\toprule
 \textbf{Sampling strategy} & \textbf{Timestep (train)}   &\textbf{Success rate} & \textbf{Model}\\\midrule
\multirow{2}{*}{Original} &Randomized  &81.4 \\\cmidrule{2-3}
 &None  &77 & \multirow{4}{*}{RVT-2}   \\\cmidrule{1-3}
\multirow{4}{*}{\makecell{Keyframes Only \\ (Ours)}} &None &81.6   &\\\cmidrule{2-3}
 &Correct &74.1  & \\\cmidrule{4-4}\cmidrule{2-3}
 &None &81.6 & \multirow{2}{*}{ARP}  \\\cmidrule{2-3}
 &Correct  &77.8 & \\
\bottomrule
\end{tabular}}
\end{table}

\begin{figure*}
    \centering
    \includegraphics[width=\linewidth]{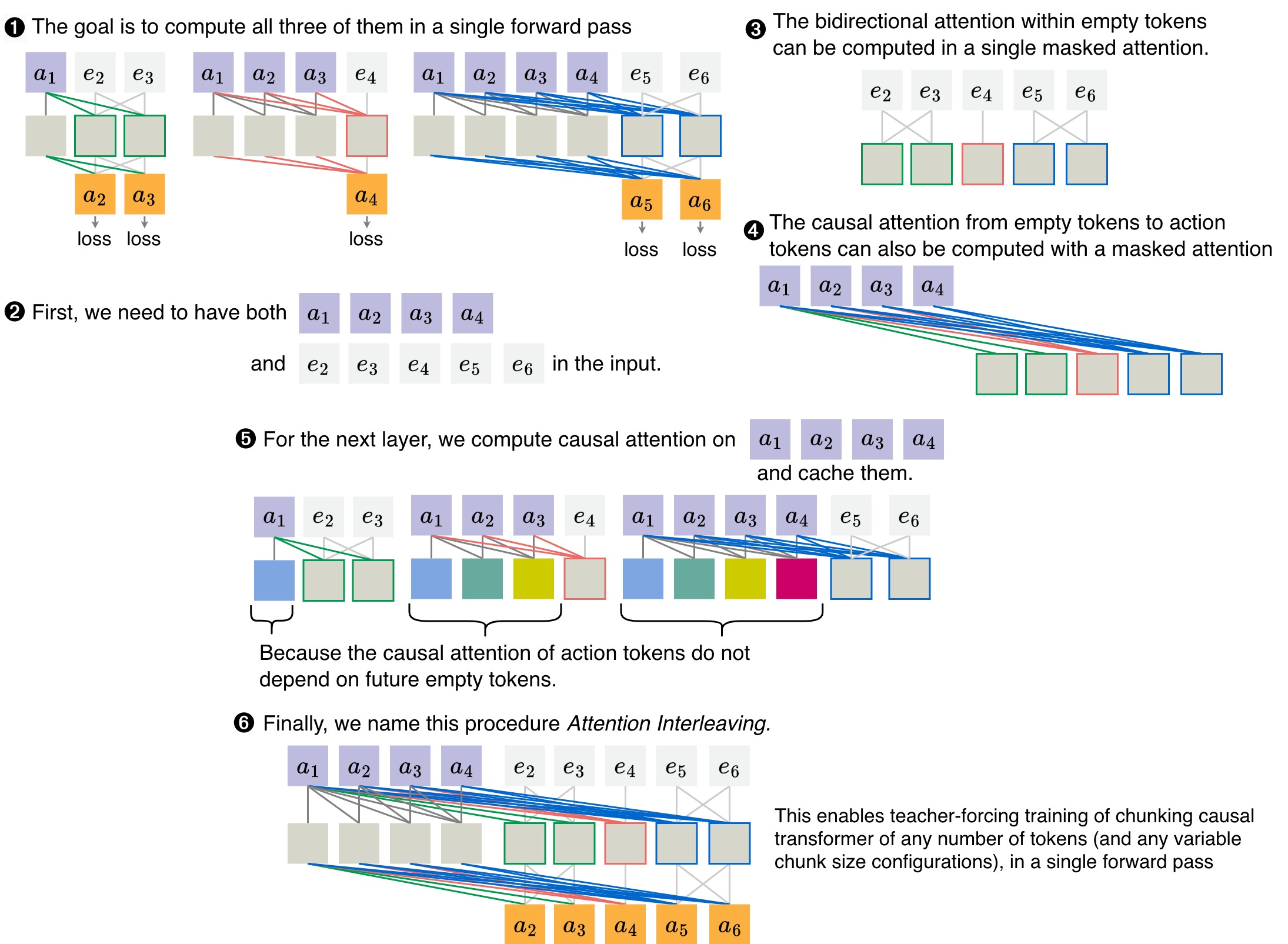}
    \caption{\textbf{Step-by-Step Explanation of Attention Interleaving.} We provide a video version of this figure \texttt{Video/attention-interleaving-tour.mp4} in the supplementary.}
    \label{fig:attn-interleave-video}
\end{figure*}

\finalrev{It is important to note that the potentially flawed design of this data-loader originates from the early work, C2FARM~\cite{james2022coarse}, rather than RVT-2. This legacy design was subsequently adopted by PerAct, RVT, and RVT-2 to ensure fair comparisons by maintaining consistency in the training data distribution.} We hope our released implementation of this simplified data-loader and ARP can be helpful for future research in RLBench or similar robotics environments.

\end{document}